\documentclass[10pt,twocolumn,letterpaper]{article}

\usepackage[cvprfinal]{cvpr}      
\usepackage{graphicx}
\usepackage{amsmath}
\usepackage{amssymb}
\usepackage{booktabs}
\usepackage{cite}
\usepackage{amsmath}
\usepackage[numbers]{natbib}
\usepackage{bbding}
\usepackage[pagebackref,breaklinks,colorlinks]{hyperref}
\usepackage[capitalize]{cleveref}
\crefname{section}{Sec.}{Secs.}
\Crefname{section}{Section}{Sections}
\Crefname{table}{Table}{Tables}
\crefname{table}{Tab.}{Tabs.}

\usepackage{amssymb}
\usepackage[ruled]{algorithm2e}
\usepackage{array}
\usepackage{enumitem}
\usepackage[bb=boondox,scr=boondoxo,scrscaled=1.1]{mathalfa}
\usepackage{mathtools}
\usepackage{subcaption}
\usepackage{wasysym}
\usepackage[dvipsnames]{xcolor}
\usepackage{lmodern}
\usepackage{siunitx}
\usepackage{booktabs}
\usepackage{etoolbox}

\renewrobustcmd{\bfseries}{\fontseries{b}\selectfont}
\renewrobustcmd{\boldmath}{}
\newrobustcmd{\B}{\bfseries}
\def\sota{state-of-the-art }

\def\cvprPaperID{9998} 
\def\confName{CVPR}
\def\confYear{2022}

\begin{document}

\title{Shuffle Augmentation of Features from Unlabeled Data for\\Unsupervised Domain Adaptation}

\author{Changwei Xu$^*$\\
Xpeng motors\\
{\tt\small chwxu@outlook.com}
\and
Jianfei Yang\thanks{These authors contributed equally.}\\
Nanyang Technological University\\
{\tt\small yang0478@e.ntu.edu.sg}
\and
Haoran Tang\\
University of Pennsylvania\\
{\tt\small thr99@seas.upenn.edu}
\and
Han Zou\\
Microsoft\\
{\tt\small enthalpyzou@gmail.com}
\and
Cheng Lu, Tianshuo Zhang\thanks{Work done while at Xpeng motors.}\\
Xpeng motors\\
{\tt\small luc@xiaopeng.com},\ {\tt\small tonyzhang2035@gmail.com}
}
\maketitle

\begin{abstract}
Unsupervised Domain Adaptation (UDA), a branch of transfer learning where labels for target samples are unavailable,
has been widely researched and developed in recent years with the help of adversarially trained models.
Although existing UDA algorithms are able to guide neural networks to extract transferable and discriminative features,
classifiers are merely trained under the supervision of labeled source data.
Given the inevitable discrepancy between source and target domains,
the classifiers can hardly be aware of the target classification boundaries.
In this paper, Shuffle Augmentation of Features (SAF), a novel UDA framework,
is proposed to address the problem by providing the classifier with supervisory signals from target feature representations.
SAF learns from the target samples, adaptively distills class-aware target features,
and implicitly guides the classifier to find comprehensive class borders.
Demonstrated by extensive experiments,
the SAF module can be integrated into any existing adversarial UDA models to achieve performance improvements.
\end{abstract}

\section{Introduction}

\begin{figure}[t]
	\centering
	\begin{subfigure}{.45\linewidth}
		\centering
		\includegraphics[width=\textwidth]{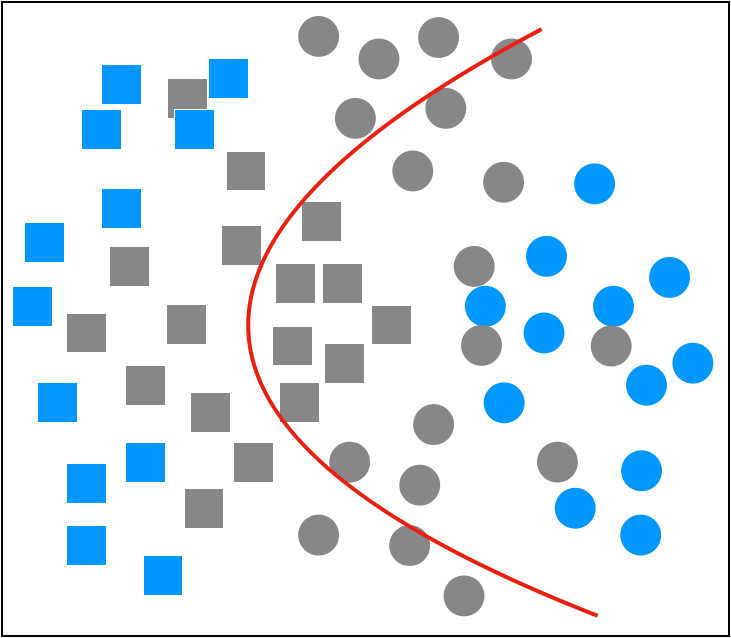}
		\caption{\small
			source only
		}
		\label{fig:intuition_a}
	\end{subfigure}
	\begin{subfigure}{.45\linewidth}
		\centering
		\includegraphics[width=\textwidth]{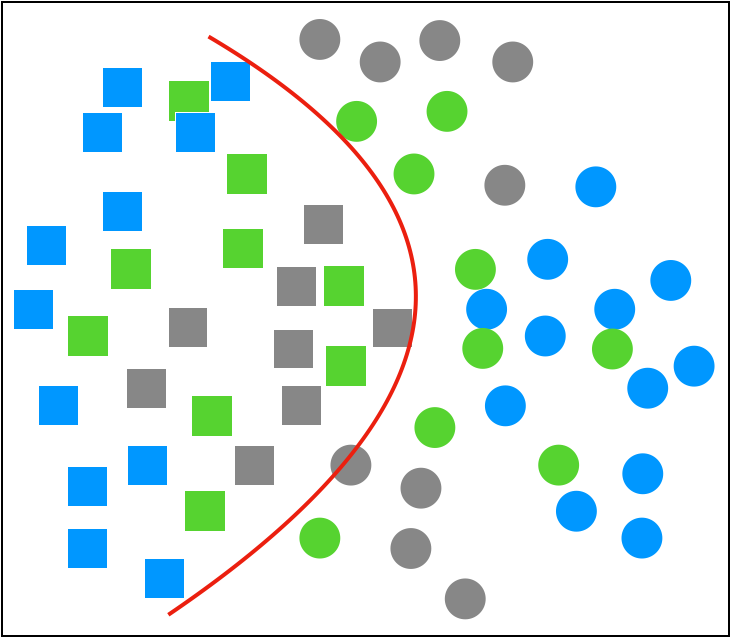}
		\caption{\small SAF assisted}
		\label{fig:intuition_b}
	\end{subfigure}
	\caption{
		\small
		Illustration of the \textit{Shuffle Augmentation of Features (SAF)} algorithm. \textbf{Best viewed in color.}
		Squares and circles indicate different classes.
		\textcolor{blue}{Blue} shapes denote source samples,
		\textcolor{gray}{gray} shapes denote unrecognized target samples,
		and \textcolor{green}{green} shapes are target samples selected by SAF.
		\textbf{(\ref{sub@fig:intuition_a})} Result of a classifier merely trained with labeled source data.
		Although the classifier learns the optimal boundary for the source distribution,
		many target samples are still misclassified due to the domain discrepancy.
		\textbf{(\ref{sub@fig:intuition_b})} Result of an SAF-improved classifier trained with both source and
		composite target features selected by SAF.
		Supervised by signals from both source and target domains,
		the classifier is able to learn a more comprehensive class boundary.
	}
	\label{fig:intuition}
\end{figure}

Transfer learning algorithms usually learn class-aware information from sufficiently labeled source data and
finetune on the target domain to transfer semantic representations.
The crux of transfer learning is the unavoidable discrepancy between the source and target domains,
leading to the essential objective of eliminating such inconsistency.
Unsupervised Domain Adaptation (UDA) is a subtopic in transfer learning where target labels are unavailable,
meaning that learning domain discrepancy between datasets is even more challenging compared with its supervised counterpart.

The Domain Adaptation (DA) problem is formally introduced by \cite{bendavid2010},
where the fundamental concepts of the discrete $\mathcal{H}\Delta\mathcal{H}$-divergence
is established.
Various methods, based on deep CNN backbones \cite{alexnet,resnet,mobilenet},
are proposed to tackle UDA tasks by reducing domain discrepancy with explicit regularizations \cite{ddc,dan,coral}.
Inspired by Generative Adversarial Network (GAN) \cite{gan},
Domain Adversarial Neural Network (DANN) \cite{dann}
extracts domain-invariant features with adversarial approach
and enlightens other adversarial UDA methods \cite{adda,jan,mcd}.
Margin Disparity Discrepancy (MDD) \cite{mdd} further generalizes $\mathcal{H}\Delta\mathcal{H}$-divergence
by providing a continuous discrepancy metric.

Although UDA training strategies have been proven to be capable of guiding CNNs to extract transferable as well as categorical features,
only labeled source features are utilized for supervised training.
In the situation shown in Figure \ref{fig:intuition_a},
features of different classes are clearly separated,
but the optimal classification boundary guided by source-only supervision misclassifies numerous target samples.
Hence, expanding the supervised training datasets with target features is necessary for further improvements.

Noisy Label (NL) learning aims to train models under the presence of incorrect labels,
which is similar to the scenario of target samples with noisy pseudo-labels in UDA.
Despite major differences between the two settings,
employing NL schemes to assist UDA training in selecting creditable target features is intuitively viable.

In this paper, we propose \textit{Shuffle Augmentation of Features} (SAF), a novel UDA framework.
Inspired by NL algorithms,
SAF learns the distribution of target features,
assigns credibility weight to each entry,
and guides the classifier to learn comprehensive class boundaries,
as illustrated in Figure \ref{fig:intuition_b}.
Moreover, SAF can be integrated into any existing adversarial UDA models,
enhancing their performances.
To summarize, our contributions are:
\begin{itemize}[topsep=2pt, itemsep=2pt, parsep=0pt]
	\item We analyze the theoretic backgrounds of the UDA problem,
	reveal the shortcomings of UDA methods with source-only supervisions,
	and discover a novel solution on the basis of cutting-edge DA theories.
	Specifically, we tackle the issues by adding target samples to supervisory signals.
	To the best of our knowledge,
	we are the first to distill class-variant information from the entire target distributions.
	\item We assimilate ideas from NL methods,
	adapt procedural details to compensate major differences between NL and UDA settings,
	and integrate their essence into the design of UDA algorithms.
	\item We propose the \textit{Shuffle Augmentation of Features} (SAF),
	which can be built upon any adversarial UDA methods.
	SAF learns from target domain distributions,
	assigns reliablility weights to pseudo-labeled samples,
	and augments the supervisory training signal.
	\item We demonstrate the superiority of SAF algorithm with extensive experiments on benchmark datasets,
	where the proposed method outperforms existing UDA models and presents the \sota performance.
\end{itemize}

\section{Related Work}

\subsection{Unsupervised Domain Adaptation}
The fundamental theory for the Domain Adaptation (DA) is formally introduced by \cite{bendavid2007,bendavid2010},
where the $\mathcal{H}\Delta\mathcal{H}$-divergence along with
the estimation formula for error boundary on the target distribution are formalized.
Based on above theories, researchers combine deep CNN \cite{rumel1985,lecun1989} backbones \cite{alexnet,resnet,mobilenet} with statistical 
measurements \cite{ddc,dan,coral,rtn,vada,mcs,afn} to shrink the gap between source and target features.
Inspired by Generative Adversarial Network (GAN) \cite{gan}, \cite{dann,dann2} proposes the Domain Adversarial Neural Network (DANN),
which is the first method employing an adversary player to implicitly align domain features.
The adversarial design of DANN further inspires numerous algorithms \cite{adda,jan,asymtri,mada,mcd,cdan,dta}.

Nevertheless, \cite{bsp} reveals that
the discriminability of the features extracted by DANN actually decreases at the same time.
\cite{mdd}
ascribes the phenomenon to the
limitations of the $\mathcal{H}\Delta\mathcal{H}$-divergence
and proposes the Margin Disparity Discrepancy (MDD),
an alternative inconsistency metric generalizable to multi-way classifiers,
to estimate the $\mathcal{H}\Delta\mathcal{H}$-divergence.

Moreover, clustering-based methods \cite{cat,impali} assign virtual labels to target data and enforce the network to follow the cluster assumption \cite{grandvalet2005}.
GAN-based methods \cite{gta,duplexgan,cccgan} augment data by directly employing GANs,
while \textit{mixup}-based methods \cite{vmt,dmrl,emixnet}
combine existing samples with the mixup \cite{mixup} technique.

\subsection{Noisy Label Learning}
The Noisy Label (NL) learning aims to train models from massive but inaccurately labeled data in real-world scenarios,
as clean datasets are usually expensive to obtain.
Traditional NL algorithms address the issue by using statistical clustering techniques
(\textit{e.g.} bagging, boosting, \textit{k}-nearest neighbor, etc.)
to determine the reliablility of each label \cite{nl_survey,nl_survey2}.
With the help of deep CNNs, the following approaches are proposed to tackle the noisy data:
assigning weights to samples w.r.t. their reliablility \cite{metacleaner,mlc,afm};
selection or creation of clean samples \cite{cleannet,selfie};
or dynamically renewing labels for mislabeled data \cite{dslfnl,dessln}.

\section{Preliminaries}\label{sec:prelim}

Given an input space $\mathcal{X}$ and a label space $\mathcal{Y}$,
a domain \cite{bendavid2010} is defined as a pair $\langle \mathcal{D}, f \rangle$,
where $\mathcal{D}$ is a distribution on $\mathcal{X}$,
and $f: \mathcal{X} \rightarrow \mathcal{Y}$ is a labeling function. In Unsupervised Domain Adaptation (UDA), there are two datasets: the labeled source dataset ($\mathcal{S}$, $\mathcal{Y}_S$),
and the unlabeled target dataset $(\mathcal{T},\varnothing)$,
drawn from different domains $\langle \mathcal{D}_S, f_S \rangle$, $\langle \mathcal{D}_T, f_T \rangle$, respectively.
Due to the discrepancy between two domains,
techniques capable of mining underlying statistic patterns from feature distributions are required.

\begin{figure*}[h]
	\begin{center}
		\includegraphics[width=0.91\textwidth]{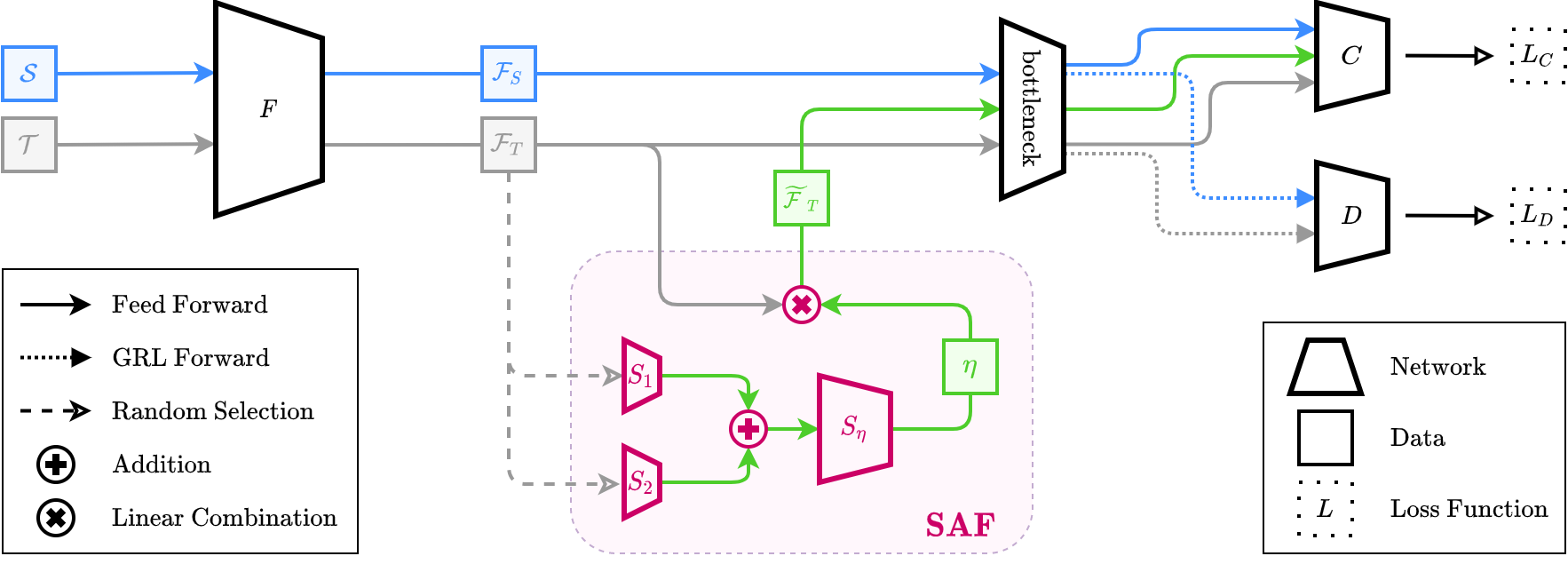}
	\end{center}
	\caption{
		\small
		The SAF framework, \textbf{best viewed in color.}
		Built upon an adversarial UDA model consisting of a feature extractor $F$, a classifier $C$, and an adversarial module $D$,
		SAF further adds an SAF-mixup module and a bottleneck layer $B$.
		SAF randomly draws pairs of target features and adaptively generates mixup weights based on target distributions.
		The combined target features $\widetilde{\mathcal{F}}_T$ and corresponding pseudo-labels are used for augmentation of supervisory signals,
		enabling the model to discover better classification boundaries for target samples.
	}
	\label{fig:structure}
\end{figure*}

Let $\mathscr{C}$ be the hypothesis space of classifiers that map from the input space $\mathcal{X}$ to $[0, 1]^{|\mathcal{Y}|}$.
For an arbitrary $C \in \mathscr{C}$, $x\in\mathcal{X}$, and $y\in\mathcal{Y}$,
denote $C_y(x)$ as the predicted probability for $x$ belonging to the class $y$,
and denote $\hat{C}(x)$ as the predicted label for $x$. \textit{Margin Disparity Discrepancy} (MDD) \cite{mdd} is a continuous metric for divergence between two domains,
which can be applied to design adversarial objectives beyond binary classifiers. Fixing the threshold $\varrho$,
the empirical discrepancy $\hat{d}_{C, \mathscr{C}}^{(\varrho)}$ for a classifier $C$ on the hypothesis space $ \mathscr{C}$ w.r.t. datasets $\mathcal{S}$,
$\mathcal{T}$ is defined as:
\begin{equation}
\hat{d}_{C, \mathscr{C}}^{(\varrho)} (\mathcal{S}, \mathcal{T}) \triangleq \\
2 \max_{C' \in \mathscr{C}} \big(
\hat{\Delta}_{\mathcal{S}}^{(\varrho)} (C, C') -
\hat{\Delta}_{\mathcal{T}}^{(\varrho)} (C, C')
\big),
\end{equation}
where $\hat{\Delta}_{\mathcal{S}}^{(\varrho)}, \hat{\Delta}_{\mathcal{T}}^{(\varrho)}$ are the \textit{empirical margin disparity}
on datasets $\mathcal{S}, \mathcal{T}$, respectively
(definitions can be found in the supplementary material).

With probability at least $1-3\delta$ $(\delta>0)$,
the risk for a classifier $C$ on the target domain $(\mathcal{D}_T, f_T)$ satisfies the following inequality:
\begin{equation}
\label{eqn:errbd2}
\begin{split}
\epsilon_{\mathcal{D}_T}(C) \le \hat{\epsilon}_{\mathcal{S}}^{(\varrho)}(C)
&+ \hat{d}_{C, \mathscr{C}}^{(\varrho)} (\mathcal{S}, \mathcal{T})
+ \lambda_{(\varrho, \mathscr{C}, \mathcal{D}_S, \mathcal{D}_T)} \\
&+ 2\sqrt{\frac{\log\frac{2}{\delta}}{2|\mathcal{S}|}} + \sqrt{\frac{\log\frac{2}{\delta}}{2|\mathcal{T}|}} \\
&+ K \Big(|\mathcal{Y}|, \frac{1}{\varrho}, \frac{1}{\sqrt{|\mathcal{S}|}}, \frac{1}{\sqrt{|\mathcal{T}|}} \Big),
\end{split}
\end{equation}
where $\hat{\epsilon}_{\mathcal{S}}^{(\varrho)}(C)$ is the empirical source risk,
$\lambda$ is a constant, and
$K$ is a term positively related to the number of classes $|\mathcal{Y}|$
and negatively related to the threshold $\varrho$ and the sizes of both source and target datasets.

However, following the definitions above,
the concept of the hypothesis class $ \mathscr{C}$ is questionable and not self-contained.
For a fixed classifier structure, the hypothesis class $ \mathscr{C}$ is invariable,
but shrinking the discrepancy $\hat{d}_{C, \mathscr{C}}^{(\varrho)} (\mathcal{S}, \mathcal{T})$ requires altering $ \mathscr{C}$.
In fact, another possible interpretation can be made to understand the theoretical backgrounds.
Guided by various supervisory signals (\textit{e.g.} loss criterions, regularizations), 
the process of optimizing CNNs can be regarded as squeezing the hypothesis class.
Supervisions aim to lead CNNs to the optimized neighborhood $ \mathscr{C}^*$
where the local discrepancy $\hat{d}_{C, \mathscr{C}^*}^{(\varrho)} (\mathcal{S}, \mathcal{T})$ is the minimum across the universe:
\begin{equation}
\mathscr{C}^* \coloneqq
\arg\min_{ \mathscr{C}' \subset \mathscr{C} }
\big[ \max_{ C \in \mathscr{C}' }
\hat{d}_{ C, \mathscr{C}' }^{(\varrho)} (\mathcal{S}, \mathcal{T})
\big].
\end{equation}

In the rest of this paper,
$ \mathscr{C}$ denotes the hypothesis neighborhood of the current classifier $C$ guided by training strategies.

\section{Shuffle Augmentation of Features}

We propose the Shuffle Augmentation of Features (SAF) algorithm for the UDA problem.

\subsection{Motivations}\label{sec:motivations}
UDA is a special branch in the field of transfer learning,
as there is no explicit clues about the target distribution
and differences between domains $\mathcal{D}_S$ and $\mathcal{D}_T$.
Existing state-of-the-art models \cite{dann2,mdd,emixnet} claim that they are capable of extracting features 
with decent transferability and discriminability from both source and target samples,
but few of these algorithms utilize the target features for the training of the classifier $C$,
yielding a suboptimal situation illustrated in Figure \ref{fig:intuition_a},
where the optimal boundary for the source distribution misclassifies numerous target samples.
Therefore, valuable target features shall be exploited to refine the classification boundaries,
as shown in Figure \ref{fig:intuition_b}.

The presence of labels is required for supervised training,
so the estimated labels $\widehat{\mathcal{Y}}_\mathcal{T}$ are required.
This introduces another intractable issue:
since the pseudo-label estimation could not be completely accurate \cite{impali,emixnet},
the network inevitably faces noisy labels.
As the model needs to learn from correct labels and diminish the negative influences from incorrect labels,
noise-robust algorithms are required to guarantee the convergence.
Intuitively, NL learning \cite{nl_survey,nl_survey2} is the desirable field where noise-robust algorithms have been widely studied and deeply developed.

\subsection{Gaps Between NL and UDA}
Although NL algorithms can effectively extract features with trustworthy labels from a noisy dataset,
it is impractical to naively transfer NL methods to UDA due to three distinctions between the two settings:
\begin{enumerate}[topsep=2pt, itemsep=2pt, parsep=0pt]
	\item In NL learning, all samples are from a unified domain distribution,
	but source and target distributions are inconsistent in the UDA setting.
	\item Popular NL benchmark datasets \cite{food101n,clothing1m,webvision} often
	contain massive amounts of samples, which are indeed favorable to pattern mining and clustering NL algorithms.
	By contrast, pervasive UDA benchmark datasets \cite{off31,offcal,offhm} offer only thousands or less samples per domain,
	where CNNs are more likely to overfit.
	\item In the NL setting, the noise ratio are usually within the range of 8\% to 38.5\% \cite{nl_survey};
	but in UDA,
	source labels are always clean and there is no guaranteed maximal noise ratio for virtually labeled target samples.
\end{enumerate}
The above differences between the two settings can potentially jeopardize the fitting of the NL training schemes,
if transferred to the circumstance of UDA without reasonable adaptations.

\subsection{Algorithm and Training Objectives}\label{sec:algo_objective}
Tailored to the characteristics of the UDA setting,
the SAF algorithm picks and incorporates the ideas of multiple NL methods \cite{metacleaner,dslfnl,afm}.
The SAF framework is built upon an adversarial UDA backbone, where three main components
(the feature generator $F$, the classifier $C$, and the adversarial module $D$) are presented.
In addition, an SAF module and a bottleneck layer $B$ are added to the SAF algorithm.
The SAF framework is shown in Figure \ref{fig:structure}.

\noindent
\textbf{Source Label Prediction Objective.}
Following the inequation \ref{eqn:errbd2} for the error boundary,
the minimization of the source classification loss is necessary for decreasing the target risk.
Since the bottleneck layer $B$ is added to regularize feature representations and takes part in prediction tasks,
the training objective of source label prediction becomes:
\begin{equation}
(F^*, B^*, C^*) \triangleq \arg \min_{F, B, C} L_C \big( [C \circ B \circ F](\mathcal{S}), \mathcal{Y_S} \big),
\end{equation}
where $L_C$ is the classification criterion, usually the cross-entropy loss.

\noindent
\textbf{Adversarial Domain Adaptation Objective.}
The domain adversarial loss $L_D$ indicates the discrepancy between feature distributions $\mathcal{F}_S$ and $\mathcal{F}_T$.
With the help of Gradient Reversal Layer (GRL) \cite{dann},
$L_D$ guides the adversary module $D$ to regularize the extractor $F$ for finding features with better transferability.

In the proposed SAF framework, $(F, B)$ and $D$ play mini-max games to attain the optimum.
In general, the domain adversarial objectives can be written as:
\begin{equation}
(F^*, B^*) \triangleq \arg \min_{F, B} \lambda_{D} L_D (F, B, D^*, C^*, \mathcal{S}, \mathcal{T}),
\end{equation}
\begin{equation}
D^* \triangleq \arg \max_{D} L_D (F^*, B^*, D, C^*, \mathcal{S}, \mathcal{T}).
\end{equation}
where $\lambda_{D}$ is the GRL weight.
The formula of $L_D$ completely depends on the design of the adversarial module $D$,
which is determined by the UDA backbone selected.

For instance, if the MDD framework \cite{mdd} were to be chosen as the backbone, $L_D$ would be:
\begin{multline}
L_D (F, B, D, C, \mathcal{S}, \mathcal{T}) \coloneqq \\
\mathbb{E}_{x\sim\mathcal{T}} \Big[ \mathcal{L}_{\text{NLL}} \Big(
1- \sigma \big( [D \circ B \circ F](x) \big)
\Big) \Big] - \\
e^{\varrho} \cdot \mathbb{E}_{x\sim\mathcal{S}} \Big[ - \mathcal{L}_{\text{NLL}} \Big(
\sigma \big( [C \circ B \circ F](x) \big)
\Big) \Big],
\end{multline}
where $\mathcal{L}_{\text{NLL}}$ is the negative log likelihood loss,
and $\sigma$ is the softmax function.

\noindent
\textbf{SAF-Supervision Objective.}
The SAF module (denoted as $M$) distills valuable target features via \textit{SAF-mixup},
a novel mixup variant, to train the classifier $C$.
As shown in Figure \ref{fig:structure}, the SAF module consists of two parallel bottlenecks $S_1, S_2$,
a weight estimator $S_{\eta}$, and two operation gates: a matrix addition gate $\bigoplus$ and a linear combination gate $\bigotimes$.

The motivation of using two independent SAF bottlenecks is straightforward:
two NNs are able to extract diversified knowledge on the selection of valuable features,
and they can learn from each other by backpropagation signals,
which provides exceptional efficacy in training.
In contrast, a single network not only lacks the diversity of a pair, but also feeds simplified information to $S_\eta$,
impairing the effectiveness of the SAF module.
The comparison is also investigated in ablation study (Table \ref{table:ablation}).

\begin{algorithm}[t]
	\small
	\LinesNumbered
	\SetAlgoLined
	\SetAlgoLongEnd
	\DontPrintSemicolon
	\SetKwInput{KwModule}{Module}
	\SetKw{KwBegin}{BEGIN:}
	\SetKw{KwEnd}{END.}
	
	\small
	\caption{SAF-mixup\label{algo:saf}}
	
	\KwModule{
		bottleneck $B$, classifier $C$, SAF bottlenecks $S_1$, $S_2$, SAF weight estimator $S_{\eta}$
	}
	\KwIn{the set of target features $\mathcal{F}_T$}
	\KwOut{augmented features ($\widetilde{\mathcal{F}}_T$, $\widetilde{\mathcal{Y}}_T$)}
	
	\KwBegin{}\;
	$\widetilde{\mathcal{F}}_T \gets \varnothing$;\: $\widetilde{\mathcal{Y}}_T \gets \varnothing$ \;
	\While(){$\mathcal{F}_T \neq \varnothing$}{
		$\phi_1, \phi_2 \gets \operatorname{RandomDrawPair} (\mathcal{F}_T)$ \;
		$\eta \gets S_{\eta}\big(S_1(\phi_1) + S_2(\phi_2)\big)$ \;
		$\tilde{\phi} \gets \eta \phi_1 + (1-\eta)\phi_2$;\;
		$\hat{y}_1 \gets C \circ B (\phi_1);\: \hat{y}_2 \gets C \circ B (\phi_2)$ \;
		$\tilde{y} \gets \eta \hat{y}_1 + (1-\eta)\hat{y}_2$ \;
		$\widetilde{\mathcal{F}}_T \gets \widetilde{\mathcal{F}}_T \cup \{\tilde{\phi}\}$;\:
		$\widetilde{\mathcal{Y}}_T \gets \widetilde{\mathcal{Y}}_T \cup \{\tilde{y}\}$ \;
	}
	\KwEnd
\end{algorithm}

In SAF-mixup, target features extracted by $F$ are randomly selected as pairs, and parallelly fed into SAF bottlenecks.
Then the sum of filtered features are forwarded into $S_{\eta}$,
where the mixup coefficient $\eta$ is determined according to the relative confidence of the paired features.
Afterwards, target features $\mathcal{F}_T$ and corresponding pseudo-labels $\widehat{\mathcal{Y}}_T$
are linearly combined with mixup coefficients to yield the augmentation dataset $(\widetilde{\mathcal{F}}_T, \widetilde{\mathcal{Y}}_T)$.
As the augmented target features are forwarded into the bottleneck $B$ and the classifier $C$,
the supervisory loss signals can be calculated and broadcasted to network structures via backpropagation.
The SAF-supervision loss is calculated using the output logits and the corresponding composite labels $\widetilde{\mathcal{Y}}_T$:
\begin{equation}
L_M(\widetilde{\mathcal{F}}_T, \widetilde{\mathcal{Y}}_T) \coloneqq
\mathcal{L}_{\text{CED}} \big( [C \circ B](\widetilde{\mathcal{F}}_T), \widetilde{\mathcal{Y}}_T \big),
\end{equation}
where $\mathcal{L}_{\text{CED}}$ is the \textit{cross-entropy divergence} (defined in the supplementary material).

Accordingly, the SAF-supervision objective can be described as the following:
\begin{equation}
(F^*, M^*, B^*, C^*) \triangleq
\arg \min_{F, M, B, C} \lambda_{M} L_M ( \widetilde{\mathcal{F}}_T, \widetilde{\mathcal{Y}}_T),
\end{equation}
where $\lambda_M$ is the SAF weight.

In SAF framework, $M$ is an adaptive Multi-Layer Perceptron (MLP) trained simultaneously with other prediction modules,
as $M$ is adjusted to yield reliable mixup coefficients. 
Since $F, B, C$ are trained with comparatively ample data samples,
beneficial mutations of $M$ will be positively amplified,
while harmful gradients of $M$ that negatively affects the fitting can be amended and rectified.

The SAF module makes up the deficiencies of UDA models with source-only supervision.
With the help of SAF, models are able to learn both source and target distributions and to perceive more accurate boundaries for target samples.

\noindent
\textbf{The Bottleneck Layer.}
As $C$ is only trained with source features, target feature fragments generated by SAF would be treated as noises
and little meaningful information about the target distributions could be obtained.
To address this issue, we propose the bottleneck $B$,
which is able to reshape $\widetilde{\mathcal{F}}_T$ to match the subsequent classifier $C$,
because $B$ learns from both source and target distributions.
The necessity of the bottleneck in the SAF training scheme is also verified quantitatively in Section \ref{sec:ablation}.

\subsection{Theoretical Insight}\label{sec:theoretical_insight}
We provide a theoretical justification of how the proposed SAF squeezes the error boundary according to the margin theories \cite{mdd}, and explains why SAF effectively seeks for novel weight-assigning strategy for noisy label learning. More details have been written in the Appendix. 
Let $C$ be the hypothesis we are interested in and denote $ \mathscr{C}$ as its hypothesis neighborhood.
Since $C$ digests feature representations $\mathcal{F}_S$ and $\mathcal{F}_T$,
the error boundary in formula \ref{eqn:errbd2} can be rewritten as:
\begin{equation}
\label{eqn:errbd3}
\begin{split}
\epsilon_{\mathcal{F}_T}(h) \le \epsilon_{\mathcal{F}_S}^{(\varrho)}(h)
&+ \hat{d}_{C, \mathscr{C}}^{(\varrho)} (\mathcal{F}_S, \mathcal{F}_T)
+ \lambda_{(\varrho, \mathscr{C}, \mathcal{D}_S, \mathcal{D}_T)} \\
&+ 2\sqrt{\frac{\log\frac{2}{\delta}}{2|\mathcal{F}_S|}} + \sqrt{\frac{\log\frac{2}{\delta}}{2|\mathcal{F}_T|}} \\
&+ K \Big(|\mathcal{Y}|, \frac{1}{\varrho}, \frac{1}{\sqrt{|\mathcal{F}_S|}}, \frac{1}{\sqrt{|\mathcal{F}_T|}} \Big).
\end{split}
\end{equation}

As explained previously, the SAF algorithm generates augmented target features and expands the supervisory dataset,
which is equivalent to enlarging the source sample space with $\widetilde{\mathcal{F}}_T$ and
increasing the source sample size by $|\widetilde{\mathcal{F}}_T|$.
Since the sum of specific terms in formula \ref{eqn:errbd3} decreases as we increase sample size of the source features $|\mathcal{F}_S|$,
the following inequality holds:
\begin{multline}
2 \sqrt{\frac{\log\frac{2}{\delta}}{2|\mathcal{F}_S \cup \widetilde{\mathcal{F}}_T|}} +
K \Big(|\mathcal{Y}|, \frac{1}{\varrho}, \frac{1}{\sqrt{|\mathcal{F}_S \cup \widetilde{\mathcal{F}}_T|}}, \frac{1}{\sqrt{|\mathcal{F}_T|}} \Big) \\
< 2 \sqrt{\frac{\log\frac{2}{\delta}}{2|\mathcal{F}_S|}} +
K \Big(|\mathcal{Y}|, \frac{1}{\varrho}, \frac{1}{\sqrt{|\mathcal{F}_S|}}, \frac{1}{\sqrt{|\mathcal{F}_T|}} \Big),
\end{multline}
proving that SAF is capable of squeezing the error boundary on the basis of margin theories.

Moreover, \textit{Active Learning} theories \cite{david1996al} and
applications in deep CNNs \cite{actln1imgclsf,actln2imgclsf,actln3imgclsf} have proven that 
learning from data with uncertain labels dramatically improves the performance of deep CNNs.
Using the conditional entropy
$
H \big( C(x) \big) = \sum_{y\in\mathcal{Y}} C_y(x) \log C_y(x) \:
$
as the uncertainty criterion,
samples with ambiguous predictions shall be reused to reinforce the model fitting.
On the one hand, reliable (or low-entropy) target samples, usually close to existing source clusters,
intensify the biases from the source distribution
instead of guiding the model to transfer semantic expressions.
On the other hand, pseudo-labels for unreliable samples are more likely to be incorrect and noisy,
which also degrade the model performance.

Therefore, the shuffle mixup among all target samples is utilized to address such dilemma.
The diversity extracted from high-entropy samples is learnt by the network to fit on target distributions,
while reliable samples offer trusty information to rectify influences of noisy gradients.
The SAF module learns from the backpropagation signals to seek for the optimal weight-assigning strategy
that balances between the biases brought by the low-entropy samples and
the noises from high-entropy samples.
The visualization in Figure \ref{fig:tsne} demonstrates the decent ability of SAF to cluster noisy target samples
while simultaneously maintaining clear class boundaries.

\begin{table*}[!t]
	\small
	\centering
	\scalebox{0.95}{
		\begin{tabular}{l | *{6}{r} | c}
			\toprule
			Method                     &   A $\to$ W\:  &   D $\to$ W\:  &    W $\to$ D\:  &   A $\to$ D\:  &   D $\to$ A\:  &   W $\to$ A\:  &   Avg \\
			\hline
			Source only                &   68.4$\pm$0.2 &   96.7$\pm$0.1 &    99.3$\pm$0.1 &   68.9$\pm$0.2 &   62.5$\pm$0.3 &   60.7$\pm$0.3 &   76.1 \\
			\hline  
			DAN \cite{dan}             &   80.5$\pm$0.4 &   97.1$\pm$0.2 &    99.6$\pm$0.1 &   78.6$\pm$0.2 &   63.6$\pm$0.3 &   62.8$\pm$0.2 &   80.4 \\
			DANN \cite{dann}           &   82.0$\pm$0.4 &   96.9$\pm$0.2 &    99.1$\pm$0.1 &   79.7$\pm$0.4 &   68.2$\pm$0.4 &   67.4$\pm$0.5 &   82.2 \\
			ADDA \cite{adda}           &   86.2$\pm$0.5 &   96.2$\pm$0.3 &    98.4$\pm$0.3 &   77.8$\pm$0.3 &   69.5$\pm$0.4 &   68.9$\pm$0.5 &   82.9 \\
			JAN \cite{jan}             &   86.0$\pm$0.4 &   96.7$\pm$0.3 &    99.7$\pm$0.1 &   85.1$\pm$0.4 &   69.2$\pm$0.3 &   70.7$\pm$0.5 &   84.6 \\
			MADA \cite{mada}           &   90.0$\pm$0.1 &   97.4$\pm$0.1 &    99.6$\pm$0.1 &   87.8$\pm$0.2 &   70.3$\pm$0.3 &   66.4$\pm$0.3 &   85.2 \\
			GTA \cite{gta}             &   89.5$\pm$0.5 &   97.9$\pm$0.3 &    99.8$\pm$0.4 &   87.7$\pm$0.5 &   72.8$\pm$0.3 &   71.4$\pm$0.4 &   86.5 \\
			MCD \cite{mcd}             &   89.6$\pm$0.2 &   98.5$\pm$0.1 & \B100.0$\pm$0.0 &   91.3$\pm$0.2 &   69.6$\pm$0.1 &   70.8$\pm$0.3 &   86.6 \\
			CDAN \cite{cdan}           &   94.1$\pm$0.1 &   98.6$\pm$0.1 & \B100.0$\pm$0.0 &   92.9$\pm$0.2 &   71.0$\pm$0.3 &   69.3$\pm$0.3 &   87.7 \\
			BSP \cite{bsp}             &   93.3$\pm$0.2 &   98.2$\pm$0.2 & \B100.0$\pm$0.0 &   93.0$\pm$0.2 &   73.6$\pm$0.3 &   72.6$\pm$0.3 &   88.5 \\
			CAT \cite{cat}             &   94.4$\pm$0.1 &   98.0$\pm$0.2 & \B100.0$\pm$0.0 &   90.8$\pm$1.8 &   72.2$\pm$0.2 &   70.2$\pm$0.1 &   87.6 \\
			SymNets \cite{symnets1}    &   90.8$\pm$0.1 &   98.8$\pm$0.3 & \B100.0$\pm$0.0 &   93.9$\pm$0.5 &   74.6$\pm$0.6 &   72.5$\pm$0.5 &   88.4 \\
			ImA \cite{impali}          &   90.3$\pm$0.2 &   98.7$\pm$0.1 &    99.8$\pm$0.0 &   92.1$\pm$0.5 &   75.3$\pm$0.2 &   74.9$\pm$0.3 &   88.8 \\
			CCC-GAN \cite{cccgan}      &   93.7$\pm$0.2 &   98.5$\pm$0.1 &    99.8$\pm$0.2 &   92.7$\pm$0.4 &   75.3$\pm$0.5 & \B77.8$\pm$0.1 &   89.6 \\
			E-MixNet \cite{emixnet}    &   93.0$\pm$0.3 &   99.0$\pm$0.1 & \B100.0$\pm$0.0 & \B95.6$\pm$0.2 & \B78.9$\pm$0.5 &   74.7$\pm$0.7 &   90.2 \\
			\hline
			MDD \cite{mdd}             &   94.5$\pm$0.3 &   98.4$\pm$0.1 & \B100.0$\pm$0.0 &   93.5$\pm$0.2 &   74.6$\pm$0.3 &   72.2$\pm$0.1 &   88.9 \\
			\textbf{MDD+SAF}           & \B96.7$\pm$0.3 & \B99.3$\pm$0.2 & \B100.0$\pm$0.0 &   94.4$\pm$0.2 &   77.2$\pm$0.1 &   75.7$\pm$0.3 & \B90.5 \\
			\bottomrule
	\end{tabular}}
	\caption{\small Accuracy (\%) on Office-31 for UDA (ResNet-50).}
	\label{table:off31}
\end{table*}

\subsection{Comparisons to Other Mixup Methods}\label{sec:comparison_mixup}

\textit{Virtual Mixup Training} (VMT) \cite{vmt}
employs mixup on raw inputs
to impose Local Lipschitzness (LL) constraint across the input space,
for the enhancement of classifier training.
\textit{Dual Mixup Regularized Learning} (DMRL) \cite{dmrl} 
further improves VMT by applying the mixup regularization on the domain classifier.
Nevertheless, neither VMT nor DMRL distills classification-related information from target samples while pursuing LL,
leaving the condition shown in Figure \ref{fig:intuition_a} unsolved.
\textit{E-MixNet} \cite{emixnet}
mixes each reliable target sample with a distant source sample.
However, as E-MixNet utilizes only target entries close to the existing source clusters,
it disobeys the principle of active learning,
and is unable to learn comprehensive class-variant information from the whole target distribution.

By comparison, only SAF can extract class-variant information from the entire target space
among all mixup-based UDA methods.
As the above models combine samples with random numbers (VMT, DMRL) or constants (E-MixNet),
SAF adaptively determines the weights using neural networks.
Our experimental results (Table \ref{table:ablation}) show that the performance of the SAF framework will be impaired
if the mixup module is replaced by weight generators of above methods.
This fact demonstrates that
SAF is indeed an unique and irreplaceable structure distinguished from any other existing mixup-based UDA models.

\begin{table*}[!t]
	\setlength{\tabcolsep}{1.9pt}
	\small
	\centering
	\scalebox{0.9}{
		\begin{tabular}{l | *{12}{c} | c}
			\toprule
			Method                  &  Ar$\to$Cl & Ar$\to$Pr  & Ar$\to$Rw  & Cl$\to$Ar  & Cl$\to$Pr  & Cl$\to$Rw  & Pr$\to$Ar  & Pr$\to$Cl  & Pr$\to$Rw  & Rw$\to$Ar  & Rw$\to$Cl  & Rw$\to$Pr  &   Avg  \\
			\hline  
			Source only             &   34.9 &   50.0 &   58.0 &   37.4 &   41.9 &   46.2 &   38.5 &   31.2 &   60.4 &   53.9 &   41.2 &   59.9 &   46.1 \\
			\hline
			DAN \cite{dan}          &   43.6 &   57.0 &   67.9 &   45.8 &   56.5 &   60.4 &   44.0 &   43.6 &   67.7 &   63.1 &   51.5 &   74.3 &   56.3 \\
			DANN \cite{dann}        &   45.6 &   59.3 &   70.1 &   47.0 &   58.5 &   60.9 &   46.1 &   43.7 &   68.5 &   63.2 &   51.8 &   76.8 &   57.6 \\
			JAN \cite{jan}          &   45.9 &   61.2 &   68.9 &   50.4 &   59.7 &   61.0 &   45.8 &   43.4 &   70.3 &   63.9 &   52.4 &   76.8 &   58.3 \\
			CDAN \cite{cdan}        &   50.7 &   70.6 &   76.0 &   57.6 &   70.0 &   70.0 &   57.4 &   50.9 &   77.3 &   70.9 &   56.7 &   81.6 &   65.8 \\
			BSP \cite{bsp}          &   52.0 &   68.6 &   76.1 &   58.0 &   70.3 &   70.2 &   58.6 &   50.2 &   77.6 &   72.2 &   59.3 &   81.9 &   66.3 \\
			ImA \cite{impali}       &   56.2 &   77.9 &   79.2 &   64.4 &   73.1 &   74.4 &   64.2 &   54.2 &   79.9 &   71.2 &   58.1 &   83.1 &   69.5 \\
			E-MixNet \cite{emixnet} & \B57.7 &   76.6 &   79.8 &   63.6 &   74.1 &   75.0 &   63.4 & \B56.4 &   79.7 &   72.8 & \B62.4 & \B85.5 &   70.6 \\
			\hline
			MDD \cite{mdd}          &   54.9 &   73.7 &   77.8 &   60.0 &   71.4 &   71.8 &   61.2 &   53.6 &   78.1 &   72.5 &   60.2 &   82.3 &   68.1 \\
			\textbf{MDD+SAF}        &   56.6 & \B79.8 & \B82.7 & \B68.2 & \B76.6 & \B77.5 & \B65.4 &   55.8 & \B82.5 & \B74.0 &   62.3 &   84.8 & \B72.2 \\
			\bottomrule
	\end{tabular}}
	\caption{\small Accuracy (\%) on Office-Home for UDA (ResNet-50).}
	\label{table:offhm}
\end{table*}

\section{Experiments}

The proposed SAF framework is evaluated on three benchmark datasets against existing state-of-the-art UDA methods.
The source code of SAF is available at the appendix.

\subsection{Setup}

\noindent
\textbf{Datasets.}
We evaluate our model on three popular DA benchmark datasets: Office-31 \cite{off31}, Office-Home \cite{offhm}, and VisDA2017 \cite{visda17}.
Office-31 has three domains (\textbf{A}mazon, \textbf{D}SLR, \textbf{W}ebcam) of 31 unbalanced classes,
containing 4,652 images.
Office-Home has four visually distinct domains (\textbf{Ar}t, \textbf{Cl}ipart, \textbf{Pr}oduct, \textbf{R}eal \textbf{w}orld) with 65 classes,
providing 15,500 images in total.
VisDA2017 is an abundant dataset with 12 classes in two completely different branches where
the training domain includes over 150k images of synthetic renderings of 3D models,
and the validation domain has approximately 55k real-world images.


\noindent
\textbf{Implementation.}
The SAF bottleneck structure is FC-ReLU, and the SAF weight estimator is FC-Sigmoid.
Other implementation details can be found in the supplementary material.
The SGD \cite{sgd} with Nesterov \cite{nesterov} momentum 0.9 is used as the optimizer.
Initially set to $0.004$, 
the learning rate of $B,C,M$ is $10$ times to that of pre-trained $F$.

\noindent
\textbf{Baselines.}
We compare SAF algorithm with the \sota UDA models,
\textit{e.g.} DANN \cite{dann}, CDAN \cite{cdan}, BSP \cite{bsp}, CCC-GAN \cite{cccgan}, etc.
Other mixup-based methods (VMT \cite{vmt}, DMRL \cite{dmrl}, E-MixNet \cite{emixnet}) are also chosen as baselines,
but only the latest E-MixNet is displayed because other early methods neither have results on benchmarks we choose
nor provide reproducible codes.
Moreover, as the experimental SAF model is implemented on MDD \cite{mdd} for optimized performance,
models built upon MDD (ImA \cite{impali}, E-MixNet \cite{emixnet}) are also involved for comparison. The commonly used full-training protocol \cite{dann} is employed for all experiments.
ResNet-50 pre-trained on the ImageNet is adopted as the feature extractor, and MDD \cite{mdd} chosen as the adversarial backbone to implement the SAF algorithm. All experiments are repeated for five times with different random seeds, reporting the average results.

\subsection{Results}

The results on Office-31 tasks are shown in Table \ref{table:off31}.
The MDD+SAF framework attains the highest accuracies of three tasks and the highest average accuracy among all contestants.
As for Office-Home results displayed in Table \ref{table:offhm},
MDD+SAF outperforms all baselines on 8 out of 12 tasks and also achieves the highest average accuracy.
Table \ref{table:visda17} presents the accuracies for MDD+SAF and multiple baseline models on the VisDA2017 benchmark
and again, our proposed model achieves the overall best performance.

Notice that SAF module improves the performance of MDD on every benchmark tasks,
even on tasks with poor source-only accuracies (\textit{e.g.} Cl$\to$Ar, Pr$\to$Ar, Ar$\to$Cl tasks in Office-Home),
In addition, SAF surpasses all other MDD-based models (ImA, E-MixNet) on average accuracies.

\begin{figure}[t]
	\centering
	\small
	\begin{minipage}{\linewidth}
		\begin{subfigure}{0.49\linewidth}
			\centering
			\includegraphics[width=\textwidth]{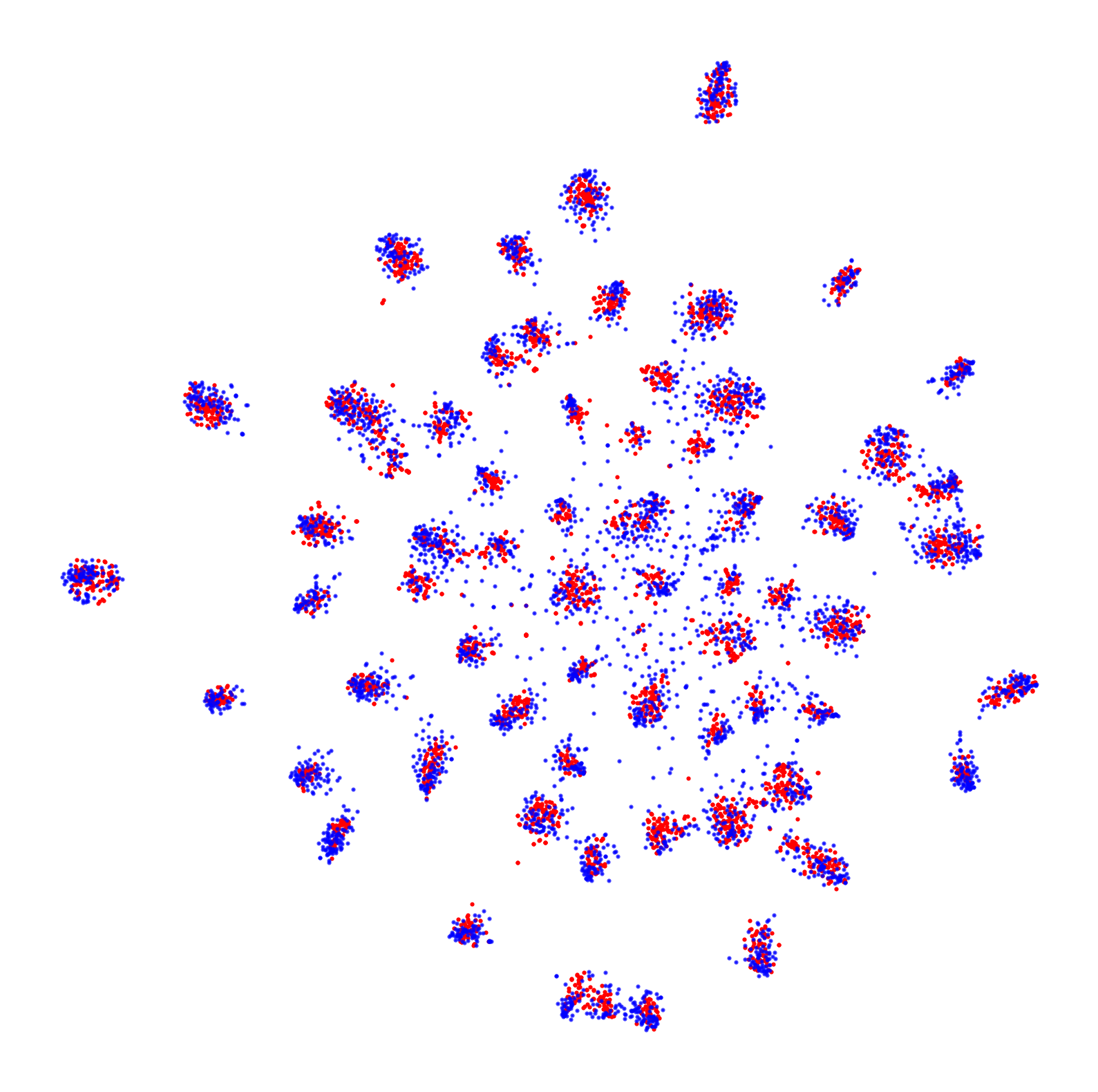}
			\caption{\small MDD: Cl$\to$Rw}
			\label{fig:tsne_mdd_ohcr}
		\end{subfigure}
		\begin{subfigure}{0.49\linewidth}
			\centering
			\includegraphics[width=\textwidth]{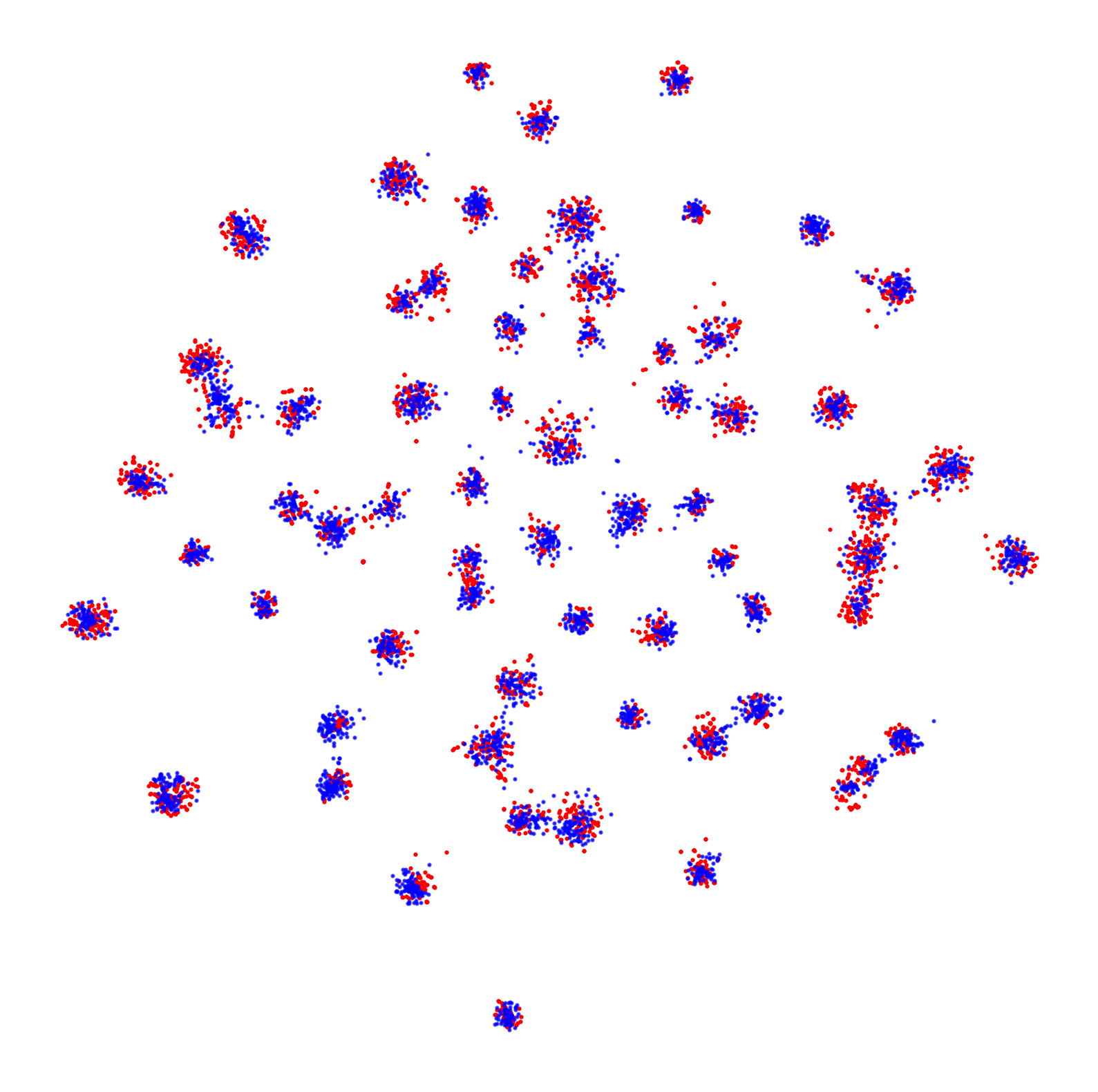}
			\caption{\small MDD+SAF: Cl$\to$Rw}
			\label{fig:tsne_saf_ohcr}
		\end{subfigure}
	\end{minipage}
	
	\begin{minipage}{\linewidth}
		\centering
		\begin{subfigure}{0.49\linewidth}
			\centering
			\includegraphics[width=\textwidth]{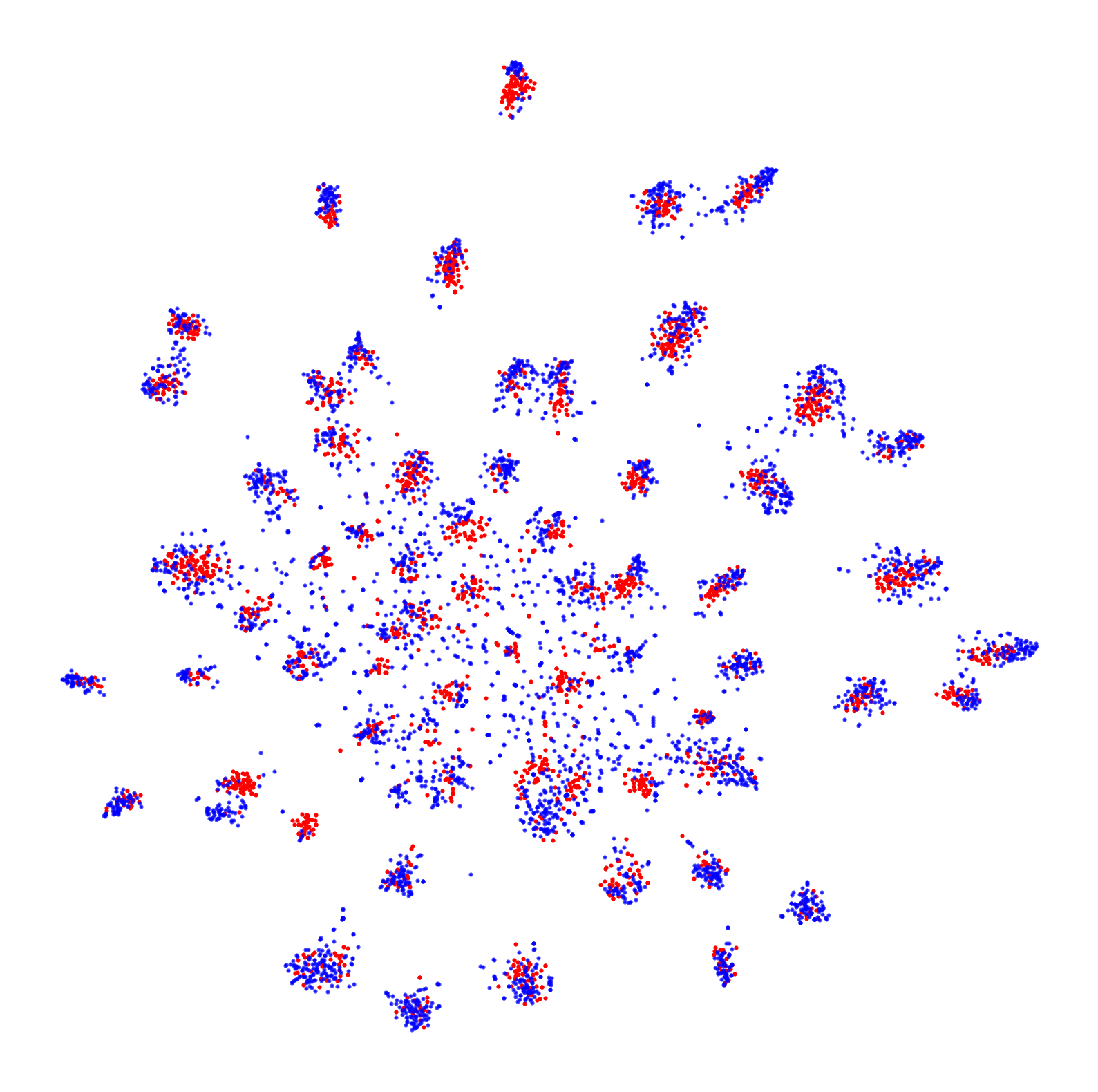}
			\caption{\small MDD: Ar$\to$Pr}
			\label{fig:tsne_mdd_ohap}
		\end{subfigure}
		\begin{subfigure}{0.49\linewidth}
			\centering
			\includegraphics[width=\textwidth]{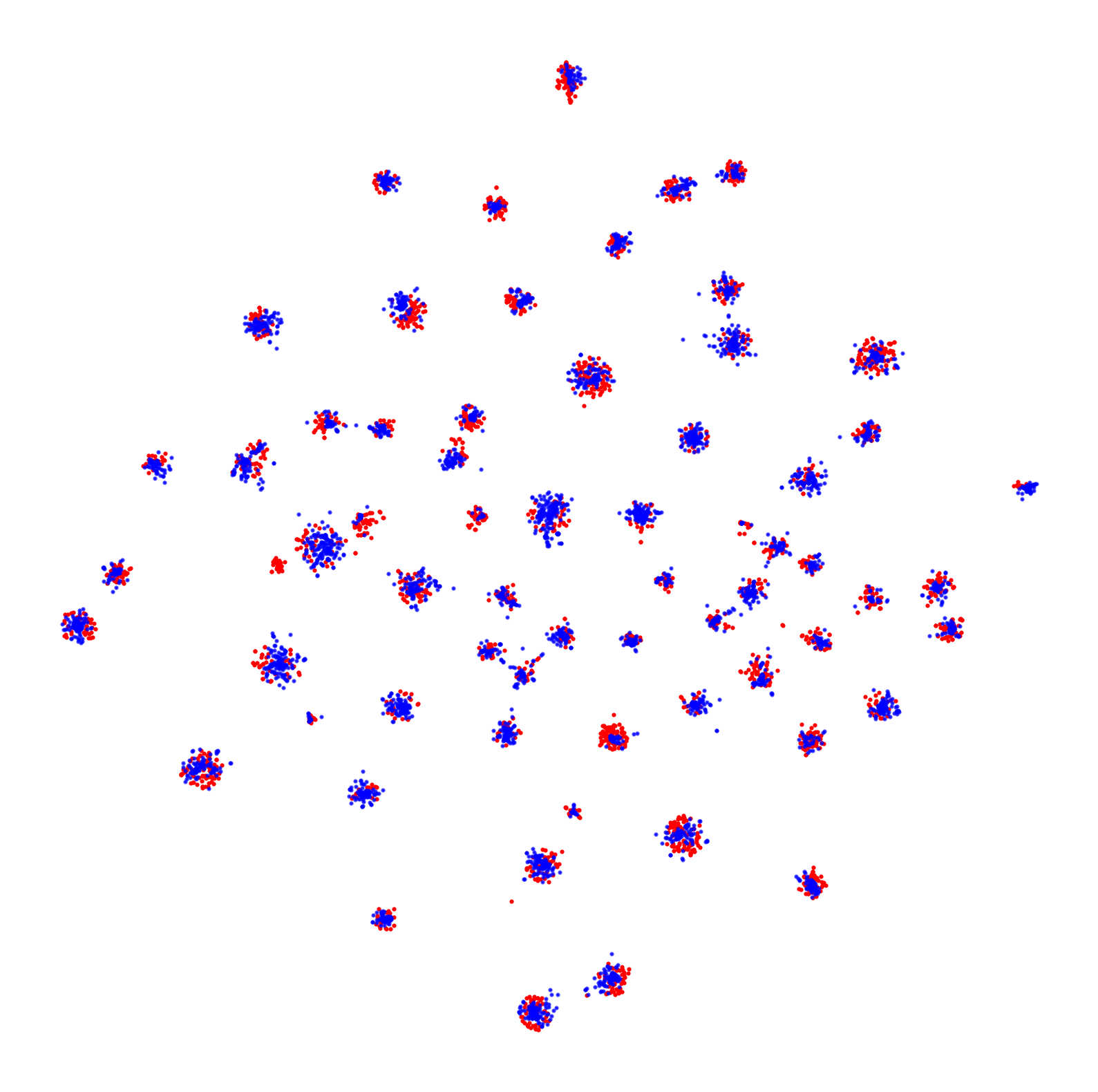}
			\caption{\small MDD+SAF: Ar$\to$Pr}
			\label{fig:tsne_saf_ohap}
		\end{subfigure}
	\end{minipage}
	\caption{\small
		t-SNE visualizations of features extracted from Office-Home tasks.
		\textbf{Best viewed in color.}
		\textcolor{red}{Red} and \textcolor{blue}{blue} dots represent source and target feature representations, respectively.
		The high-resolution version can be found in the supplementary material.
	}
	\label{fig:tsne}
\end{figure}

The t-SNE visualization \cite{tsne} shown in Figure \ref{fig:tsne},
where feature representations extracted from both domains of two Office-Home tasks are clustered via t-SNE algorithm,
strongly demonstrates the classification improvement brought by the SAF module.

\subsection{Ablation Study and Analytics}\label{sec:ablation}

\noindent\textbf{Impact of the Bottleneck.}
As mentioned in Section \ref{sec:algo_objective},
the bottleneck layer $B$ functions as a regularizer for the output of SAF to be better recognized by $C$.
The necessity of $B$ is investigated through experiments on representative tasks of Office-Home.
To keep the model capacity unchanged, $B$ needs to be preserved in place,
so the input samples are forwarded in the order $F$$\to$$B$$\to$$M$$\to$$C$.
As shown in the ``no bottleneck layer'' entry of Table \ref{table:ablation},
the removal of the bottleneck weakens the model performance.

\noindent\textbf{Impact of Adaptive SAF-mixup.}
In Section \ref{sec:comparison_mixup} we theoretically claim that
SAF-mixup is a special mixup technique different from others.
In this experiment, the SAF-mixup module is replaced by
\textbf{(1)} a beta random variable $\mathit{Beta}(0.2, 0.2)$ (used by \cite{mixup,dmrl}), and
\textbf{(2)} a constant $0.6$ (used by \cite{emixnet}).
Based on the comparison results, the modifications severely impair the accuracies,
demonstrating the significance of the adaptive SAF module.

\noindent\textbf{Impact of Two SAF Bottlenecks.}
To examine the necessity of using two separated SAF bottlenecks,
a modified SAF model with a single SAF bottleneck,
and another version with four SAF bottlenecks are tested.
The study result shows that either increasing or decreasing the number of bottlenecks will harm the model performance,
indicating that using two SAF bottlenecks is the optimal choice.

\begin{table}[t]
	\newcolumntype{L}[1]{>{\raggedright\let\newline\\\arraybackslash\hspace{0pt}}m{#1}}
	\newcolumntype{R}[1]{>{\raggedleft\let\newline\\\arraybackslash\hspace{0pt}}m{#1}}
	\newcolumntype{C}[1]{>{\centering\arraybackslash}p{#1}}
	
	\centering
	\small
	\begin{minipage}{\linewidth}
		\centering
		\scalebox{1}{
			\begin{tabular}{ l C{0.35\linewidth} }
				\toprule
				Method & Accuracy (\%) \\
				\midrule
				JAN \cite{jan}       &   61.6 \\
				GTA \cite{gta}       &   69.5 \\
				MCD \cite{mcd}       &   69.8 \\
				CDAN \cite{cdan}     &   70.0 \\
				ImA \cite{impali}    &   75.8 \\
				\midrule
				MDD \cite{mdd}       &   74.6 \\
				\textbf{MDD+SAF}     & \B77.0 \\
				\bottomrule
			\end{tabular}
		}
		\caption{\small
			VisDA2017 Accuracy for UDA (ResNet-50)
		}
		\label{table:visda17}
	\end{minipage}
\end{table}

\begin{table}[t]
	\begin{minipage}{\linewidth}
		\small
		\centering
		\scalebox{1.0}{
			\begin{tabular}{ l c c c }
				\toprule
				Modification                     &  Ar$\to$Pr  &  Pr$\to$Ar \\
				\midrule
				MDD \cite{mdd}                   &   73.7  &   61.2 \\
				no bottleneck layer              &   75.7  &   61.6 \\
				beta random variable $\eta$      &   73.7  &   57.6 \\
				constant $\eta$                  &   75.3  &   64.0 \\
				only one SAF bottleneck          &   76.0  &   62.3 \\
				four SAF bottlenecks             &   77.3  &   64.0 \\
				feed source into SAF             &   76.4  &   62.6 \\
				only mix uncertain samples       &   73.7  &   60.7 \\
				only mix certain samples         &   76.2  &   65.4 \\
				\textbf{MDD+SAF}                 & \B79.8  & \B65.4 \\
				\bottomrule
			\end{tabular}
		}
		\caption{\small
			The impact of different modifications to the SAF.
		}
		\label{table:ablation}
	\end{minipage}
	
	\small
	\centering
	\begin{minipage}{\linewidth}
		\vspace*{10pt}  
		\centering
		\scalebox{1.0}{
			\begin{tabular}{ l c c c c }
				\toprule
				Method               &   A$\to$D  &   A$\to$W  &   D$\to$W  &   Avg \\
				\midrule
				DANN \cite{dann}     &   79.7 &   82.0 &   96.9 &   82.2 \\
				\textbf{DANN+SAF}    & \B86.0 & \B88.8 & \B99.0 & \B84.5 \\
				\midrule
				CDAN \cite{cdan}     &   92.9 &   94.1 &   98.6 &   87.7 \\
				\textbf{CDAN+SAF}    & \B94.6 & \B95.1 & \B99.3 & \B89.7 \\
				\bottomrule
		\end{tabular}}
		\caption{\small
			Accuracy (\%) of SAF-improved adversarial methods (i.e. DANN \cite{dann} and CDAN \cite{cdan}).
		}
		\label{table:dann_saf}
	\end{minipage}
\end{table}

\noindent\textbf{Impact of SAF on Source Data.}
As only target features are fed into SAF,
it is natural to wonder whether forwarding the source features into SAF benefits the model.
The ``feed source into SAF'' entry in Table \ref{table:ablation} shows the result and the answer is negative.
Since the structure of SAF is relatively simple, its low model capacity disallows itself from fitting on both domains.
When SAF learns from both source and target distributions,
it loses its speciality on the target domain to compensate the fitting on the source.

\noindent\textbf{Impact of Sample Selection.}
According to active learning \cite{david1996al},
all target samples, with certain or uncertain predictions, are involved in the SAF-mixup
to assimilate diversity of high-entropy samples and stability of low-entropy samples.
The SAF module can adaptively balance the biases of source distribution and the noises brought by uncertain samples.
To verify the impact of sample selection,
we design experiments where mixup samples are filtered based on their conditional entropy with an empirical threshold.
As shown in Table \ref{table:ablation},
mixing only uncertain samples or excluding them can harm the model performance.

\noindent\textbf{Generalizability of SAF.}
The effectiveness of SAF on MDD has been verified in standard experiments.
In addition, since SAF can be plugged into arbitrary adversarial UDA models,
the general effectiveness of SAF on all eligible backbones becomes a desirable property.
Therefore, we further combine SAF with DANN \cite{dann} and CDAN \cite{cdan} to test the generalizability.
As shown in Table \ref{table:dann_saf},
the SAF plug-in strongly boosts the accuracies of DANN and CDAN backbone on Office-31 tasks,
which confirms its generalizability.

\section{Conclusion}

In this paper, a novel SAF algorithm is derived from the cutting-edge theories to address the unsupervised domain adaptation problem. The key idea of SAF is to distill classification-related information from target features to augment the training signals. Our SAF framework learns from input targets and guides the classifier to find reasonable class boundaries for the target distribution. Extensive experiments demonstrate the \sota performance and decent generalizability of SAF.

{\small
\bibliographystyle{ieee_fullname}
\bibliography{egbib}
}

\clearpage
\section*{Appendix}

\subsection{Theoretical Insight: \\Evolution of UDA Algorithms}

In this section, we introduce the path of evolution in domain adaptation theories,
along with UDA algorithms based on theoretical backgrounds over time.
To summarize, we categorize the theoretical progression of domain adaptation to two stages.
In each stage, researchers discover a specific issue in UDA,
initially tackle with traditional statistical methods (explicit techniques),
and end up with solving the problem with implicit approaches.

\subsubsection{Notations}
For a neural network specified for UDA tasks,
denote $F$, $C$, $D$ as feature extractor, classifier, and adversarial module, respectively.
The feature extractor, $F$, is a structure that distills feature representations from input images.
Usually, researchers use pretrained backbones (\textit{e.g.} AlexNet \cite{alexnet}, ResNet \cite{resnet}) for the extractor.
The classifier, $C$, is a Multi-Layer Perceptron (MLP) that assigns class labels to input features.
The adversarial module, $D$, is another MLP that is trained to regularize $F$ and $C$ through mini-max games.
The adversarial module was first proposed by \cite{dann} as a domain classifier.
$F, C, D$ are commonly presented in recent UDA frameworks,
as the adversarial design has become pervasive among UDA algorithms.

We follow the definition of \textit{domain} proposed by \cite{bendavid2010} and generalized by \cite{impali}.
Given an input space $\mathcal{X}$ and a label space $\mathcal{Y}$,
a domain $\mathscr{D}$ is a pair $\langle \mathcal{D}, f \rangle$ consisting of a distribution $\mathcal{D}$ on $\mathcal{X}$,
and a labeling function $f: \mathcal{X} \rightarrow \mathcal{Y}$.
Namely, the labeling function assigns the ground-truth label $y \in \mathcal{Y}$ to each sample $x \in \mathcal{X}$.

Let $\mathscr{D}_S = \langle \mathcal{D}_S, f_S \rangle$ denote the source domain on the input space,
with domain label $\mathcal{Z}_S = 0$.
Let ($\mathcal{S}$, $\mathcal{Y}_S$) be the set of available labeled source samples along
with empirical distribution $\widehat{\mathcal{D}}_S$.
Let $\mathcal{F}_S$ be the set of feature representations extracted from the source dataset,
with empirical feature distribution $\Phi_S$.
Moreover, denote ${P}_S$ as the set of predicted possibility vectors from the classifier based on $\mathcal{F}_S$.

The symmetric (reflected) definitions on the target domain:
$\mathscr{D}_T = \langle \mathcal{D}_T, f_T \rangle$, $\mathcal{Z}_T=1$, $(\mathcal{T}, \mathcal{Y}_T)$,
$\widehat{\mathcal{D}}_T$, $\mathcal{F}_T$, $\Phi_T$, ${P}_T$
are described in exactly the same manner as their source counterparts.

Let $\mathcal{H}$ be the hypothesis class of classifiers that maps from $\mathcal{X}$ to $\{\mathcal{Z}_S, \mathcal{Z}_T\}$.

\subsubsection{Settings}
In Unsupervised Domain Adaptation (UDA), there are two datasets:
the labeled source dataset $(\mathcal{S}, \mathcal{Y}_S)$ drawn from $\mathscr{D}_S$,
and the unlabeled target dataset $(\mathcal{T}, \varnothing)$ drawn from $\mathcal{D}_T$,
sharing the input space $\mathcal{X}$ and the label space $\mathcal{Y}$.
The crux of the UDA setting is that the discrepancy between two domains cannot be explicitly alleviated,
so methods that are capable of mining underlying statistical patterns instead of merely learning from labels,
are required.

\subsubsection{Foundations of Domain Adaptation}
We begin with the fundamental theories and basic terminologies for domain adaptation problems.
For the given domains $\mathscr{D}_S$ and $\mathscr{D}_T$,
we use the concept $\mathcal{H}$-\textit{divergence} \cite{bendavid2010} to measure their discrepancy
with respect to the hypothesis class $\mathcal{H}$:
\begin{equation}
	\label{eqn:hdiv}
	d_{\mathcal{H}}(\mathcal{D}_S, \mathcal{D}_T) \coloneqq \\
	2 \sup_{h\in\mathcal{H}} \big|
	\mathbb{P}_{\mathcal{D}_S}[h = 0] - \mathbb{P}_{\mathcal{D}_T}[h = 0]
	\big|,
\end{equation}
which can be empirically estimated if $\mathcal{H}$ is symmetric:
\begin{equation}
	\begin{split}
		\hat{d}_{\mathcal{H}}(\mathcal{S}, \mathcal{T}) \triangleq& \\
		2\Big(1-\min_{h\in\mathcal{H}}& \big[
		\frac{1}{|\mathcal{S}|} \sum_{x\in h_0} \mathbb{1}[x \in \mathcal{S}] +
		\frac{1}{|\mathcal{T}|} \sum_{x\in h_1} \mathbb{1}[x \in \mathcal{T}]
		\big]\Big),
	\end{split}
\end{equation}
where
$h_\mathcal{Z} \coloneqq \{x \in (\mathcal{S} \cup \mathcal{T}) \mid h(x) = \mathcal{Z}\}$
for $\mathcal{Z} \in \{ \mathcal{Z}_S \coloneqq 0, \mathcal{Z}_T \coloneqq 1 \}$,
and $\mathbb{1}$ is the binary indicator function.

Intuitively, $\mathcal{H}$-divergence measures the discrepancy between two domains
based on the worst domain classifier in $\mathcal{H}$.
If the worst classifier $h^- \in \mathcal{H}$ can hardly distinguish samples from two domains,
then the discrepancy between $\mathcal{S}$ and $\mathcal{T}$ would be high on $\mathcal{H}$ and \textit{vice versa}.

Based on above notations,
the $\mathcal{H}\Delta\mathcal{H}$-\textit{divergence} \cite{bendavid2010} is defined as follows:
\begin{multline}
	\label{eqn:hdhdiv}
	d_{\mathcal{H}\Delta\mathcal{H}}(\mathcal{D}_S, \mathcal{D}_T) \coloneqq \\
	2 \sup_{h, h'\in \mathcal{H}} \big|
	{\mathbb{P}}_{\mathcal{D}_S}[h \neq h'] - {\mathbb{P}}_{\mathcal{D}_T}[h \neq h']
	\big|.
\end{multline}
Following this definition,
a powerful bounding formula for the classification error on $\mathcal{D}_T$ can be derived.
For every $h \in \mathcal{H}$ on the target domain $\mathcal{D}_T$,
with probability at least $1 - \delta$ $(\delta>0)$:
\begin{equation}
	\label{eqn:errbd}
	\epsilon_{\mathcal{D}_T}(h) \le \epsilon_{\mathcal{D}_S}(h)
	+ \frac{1}{2}d_{\mathcal{H}\Delta\mathcal{H}}(\mathcal{D}_S, \mathcal{D}_T)
	+ \lambda,
\end{equation}
where $\lambda$ is a constant.

\noindent
\textbf{Hypothesis Neighborhood.}\label{sec:hypo_nbhd}
The concept of hypothesis class $\mathcal{H}$ becomes confusing in above formulas.
For a fixed CNN structure, the hypothesis space is unchangeable,
but shrinking the discrepancy term $d_{\mathcal{H}\Delta\mathcal{H}}$ requires shifting of $\mathcal{H}$.
Actually, we can regard the process of optimizing CNNs,
guided by various supervisory signals (loss criterions, regularizations), 
as shrinking the hypothesis space.
Led by supervisions, the CNNs is guided to find the optimized neighborhood,
$\mathcal{H}^*$, in the universal hypothesis class $\mathcal{H}$,
where the local $\mathcal{H}^*\Delta\mathcal{H}^*$-divergence is minimized across the universe:
\begin{equation}
	\mathcal{H}^* \coloneqq
	\arg\min_{\mathcal{H'} \subset \mathcal{H}} \big[
	d_{\mathcal{H'}\Delta\mathcal{H'}}(\mathcal{D}_S, \mathcal{D}_T)
	\big].
\end{equation}
In the rest of this document,
the symbol $\mathcal{H}$ indicates the hypothesis neighborhood of the current classifier $h$.
When designing domain adaptation algorithms, we are not interested in the universal hypothesis class,
but try to discover the best optimization strategies that drive the models to reach $\mathcal{H}^*$.

\subsubsection{Stage I: Domain Feature Alignment}
After the popularization of the CNN, researchers combine traditional statistical techniques
with deep CNNs \cite{ddc,dan,coral,rtn,vada,mcs,afn} to enhance model performance in domain adaptation tasks.
Generally, these works employ explicit methods to measure discrepancy between
feature representations $\mathcal{F}_S$ and $\mathcal{F}_T$ (the output of the feature extractor $F$),
or output probabilities ${P}_S$ and ${P}_T$ (the output of the classifier $C$).
A common practice is to design a loss term based on the calculated discrepancy and to encourage the network to align all samples.
Following formula \ref{eqn:errbd}, researchers believe that the discrepancy between $\Phi_S$ and $\Phi_T$
(the domain discrepancy among feature) greatly affects the classification accuracies.
Hence, methods in this stage strive to close the gap from two directions:
\textbf{(1)} regularize the model to ignore domain-variant features;
\textbf{(2)} push the model escape from the suboptimal neighborhoods $\overline{\mathcal{H}^*}$.
To sum up, in this stage, researchers try to distill the essence of explicit methods
for better feature alignment across different domains.

\noindent
\textbf{DANN \cite{dann}} is a revolutionary innovation that changes the situation.
In DANN, domain-variant feature representations are no longer aligned through the assistance of explicit calculations,
but implicitly achieved by training another MLP, the domain discriminator $D$.
The DANN was inspired by GAN\cite{gan}, where two independent deep neural networks, 
Generator and Discriminator, are trained together but with completely opposite goals.
The Generator aims to create fake samples mimicking real samples from random gaussian noises to fool the Discriminator,
while the Discriminator struggles to distinguish between real samples from datasets and fake samples generated by the Generator.
Abide by the idea of Discriminator, the goal of the domain discriminator $D$
is to recognize whether the features extracted by $F$ are from the source distribution or the target counterpart.
Similar to a Generator, the feature map $F$, while extracting better feature representations to aid $C$,
also needs to extract domain-invariant features shared by ${\Phi}_S$ and ${\Phi}_T$,
incapatiating $D$ from making correct predictions.

The invention of DANN signals a major change in the development of UDA algorithms:
the alignment of domain features can finally be done in an implicit NN-styled way.
Not to mention that this implicit modification outperforms existing explicit methods on popular UDA benchmarks \cite{dann2}.

\subsubsection{Stage II: Classification Feature Alignment}
As the DANN emerges, adversarial approaches become favorable among researchers,
which inspires numerous adversarial UDA methods \cite{adda,jan,asymtri,mada,mcd,cdan,dta}.
However, another cloud still obscure the sky of UDA researching:
even though domain alignment regularizations are employed,
none of the existing models can achieve the same performance on the target domain as on the source domain.

\begin{figure}
	\centering
	\begin{subfigure}[b]{0.48\linewidth}
		\centering
		\includegraphics[width=\textwidth]{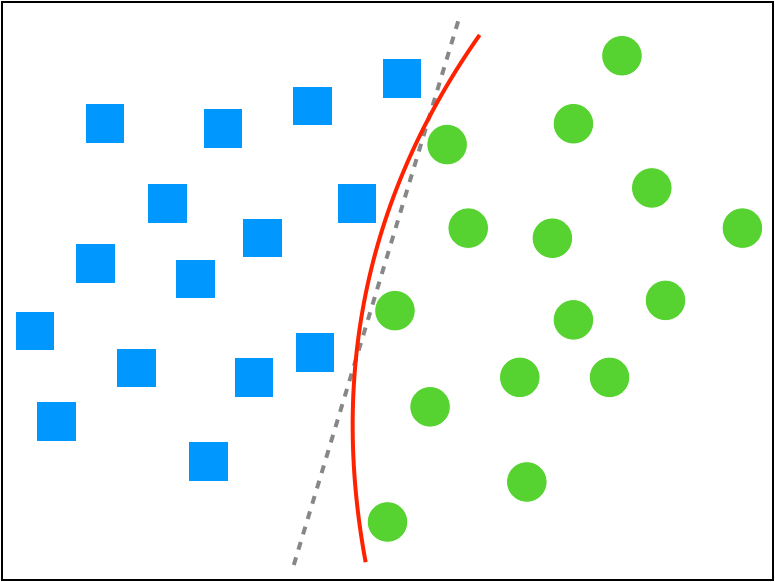}
		\caption{}
		\label{fig:discriminability_good}
	\end{subfigure}
	\begin{subfigure}[b]{0.48\linewidth}
		\centering
		\includegraphics[width=\textwidth]{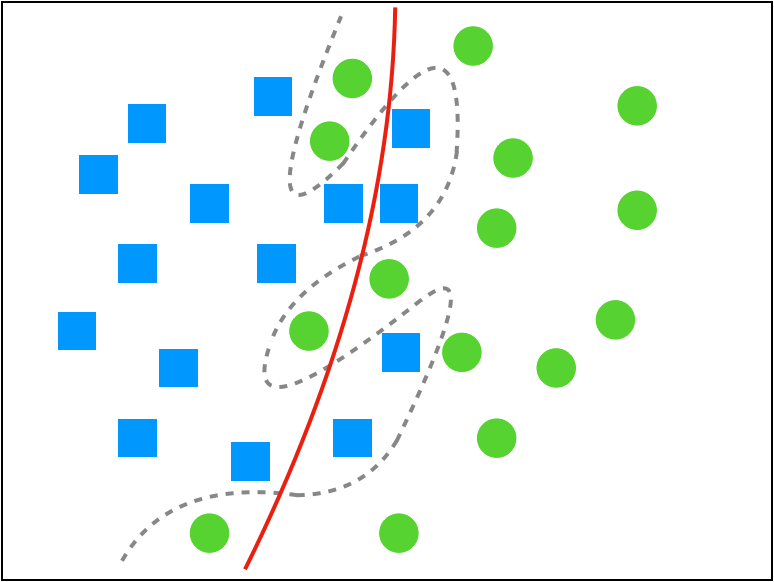}
		\caption{}
		\label{fig:discriminability_bad}
	\end{subfigure}
	\caption{
		Illustration of \textit{discriminability}.
		\textbf{Best viewed in color.}
		\textbf{(a)} shows feature distribution with decent discriminability,
		as the classification boundary can be closely estimated by the classifier.
		\textbf{(b)} shows the contrary, where the dashed line is the ground-truth class border,
		but classifier with regular capacity usually finds the solid curve as predicted broader.
	}
	\label{fig:discriminability}
\end{figure}

\noindent
\textbf{Shortcomings of DANN.}
After a seires of formal analysis and rigorous experiments,
researchers introduce two concepts that are necessary for understanding the dilemma of DANN:
transferability and discriminability \cite{bsp}.
Transferability is an attribute indicating whether the feature representations extracted by $F$ are completely shared by both domains,
\textit{i.e.} $\Phi_S = \Phi_T$.
Discriminability is another property referring to the easiness for a classifier to find clear class boundaries among input feature representations,
as shown in Figure \ref{fig:discriminability}.

Formally, the DANN structure, while boost the accuracy on UDA tasks by enhancing transferability,
actually misleads $F$ to extract feature representations with decreased discriminability \cite{bsp}.
To address this issue, multiple approaches \cite{mcd,bsp,rada} are attempted to explicitly increase
the discriminability of the features extracted by $F$ in adversarial structures.
In this stage, researchers aim to design objective functions able to lead $F$ to excavate more class-variant,
or discriminative features.

We can also understand this phenomenon by investigating the theoretical supports of DANN.
Since the $\mathcal{H}$-divergence (Equation \ref{eqn:hdiv}) and the $\mathcal{H}\Delta\mathcal{H}$-divergence (Equation \ref{eqn:hdhdiv})
employ the discrete \textit{0-1 disparity} as the risk criterion:
\begin{equation}
	\epsilon_{\mathcal{D}}(h) \coloneqq \mathbb{P}_{\mathcal{D}}\big[\mathbb{1}[h\neq f]\big],
\end{equation}
one can hardly generalize a reasonable scoring function for classifiers with more than two classes from them.
A theoretically possible but technically impractical approach is to naively create binary classifier for every pair of classes,
but this design is computationally expensive,
especially for datasets with massive classes (\textit{e.g.} ImageNet \cite{imagenet}, DomainNet \cite{domainnet}, etc.).
Hence, the binary classifier $D$ can only function as a domain regularizer and cannot boost the classification accuracy.

\noindent
\textbf{MDD \cite{mdd}.}
Some researchers regard such incompatibility as a gap between theories and algorithms in the field of UDA,
making the optimization process extremely difficult or even impossible.
To field this gap, the concept of \textit{Margin Disparity Discrepancy} (MDD) \cite{mdd} is formulated,
which is another revolutionary theoretical innovation in the UDA research.

The MDD framework is able to implicitly align classification features between two domains with adversarial approach.
MDD not only bridges the gap between existing discrete domain adaptation theories and continuous objective functions required by model training,
but also implicitizes the alignment of classification features.
In other words, MDD first employs the implicit approach in guiding the feature extractors to distill featuer representations with better discriminability.
As a result, MDD outperforms existing methods on benchmarks \cite{mdd}
and again demonstrates the superiority of implicit approaches over explicit ones.

\subsection{Definitions and Notations}
We redefine core DA concepts to adapt the MDD system.
Let $\mathscr{C}$ be the hypothesis space of classifiers that maps from the feature space $\Phi$ to $[0, 1]^{|\mathcal{Y}|}$.
For arbitrary $C \in \mathscr{C}$, $\phi\in\Phi$, and $y\in\mathcal{Y}$,
denote $C_y(\phi)$ as the predicted probability for $\phi$ belonging to the class $y$,
and denote $\hat{C}(\phi)$ as the predicted label for $\phi$.
In addition, $\mathcal{X}$ and $\Phi$ are interchangeable in this context,
as we focus on the optimization of classifiers.

\noindent
\textbf{Margin.}
Following the margin theory \cite{koltch2004}, the concept \textit{margin} \cite{mdd} of a classifier $C$ with respect to
another classifier $C'$ on a sample $x\in\mathcal{X}$ is defined as:
\begin{equation}
	\rho_C(x, C') \coloneqq \frac{1}{2} \big(C_{\hat{C'}(x)}(x) - \max_{y\neq \hat{C'}(x)}C_y(x)\big).
\end{equation}
Namely, the margin regards the label predicted by $C'$ as the exemplar label,
and measures the distance between the probability of the ``ground-truth label''
and the largest probability among the remainings.

\noindent
\textbf{Margin Loss.}
Given a task-dependent constant \textit{margin threshold} (denoted as $\varrho$),
we would like the margin criterion to satisfy the following properties:
\begin{itemize}[leftmargin=20pt, topsep=2pt, itemsep=2pt, parsep=0pt]
	\item For any input, the margin loss falls in $[0, 1]$.
	\item The loss becomes $1$ if $\rho_C(x, C')$ is negative.
	\item The loss becomes $0$ if $\rho_C(x, C')$ is larger than $\varrho$.
	\item The loss decreases as $\rho_C(x, C')$ increases in $[0, \varrho]$.
\end{itemize}
Therefore, the fundamental \textit{margin loss} \cite{mdd} of a classifier $C\in \mathscr{C}$ w.r.t.
another classifier $C'\in \mathscr{C}$
on a sample $x\in\mathcal{X}$ is defined as:
\begin{multline}
	\bar{\rho}^{(\varrho)}_{C} (x, C') \coloneqq \\
	\begin{cases}
		1                             & \text{if } \rho_C(x, C') \in (-\infty, 0) \\
		1 - \rho_C(x, C') / \varrho   & \text{if } \rho_C(x, C') \in [0, \varrho] \\
		0                             & \text{if } \rho_C(x, C') \in (\varrho, \infty)
	\end{cases}.
\end{multline}

\noindent
\textbf{Margin Prediction Risk.}
For a fixed threshold $\varrho$, define the \textit{margin prediction risk}
for a classifier $C$ on a domain $\mathscr{D} = \langle \mathcal{D}, f \rangle$ as the following:
\begin{equation}
	\epsilon_{\mathcal{D}}^{(\varrho)} (C) \coloneqq
	\mathbb{E}_{x\sim\mathcal{D}} \big[ \bar{\rho}^{(\varrho)}_{C}(x, f) \big] .
\end{equation}
The \textit{empirical margin prediction risk} on a dataset $\mathcal{U}$ with
ground-truth labeling function $f_{\mathcal{U}}$ can be calculated by:
\begin{equation}
	\hat{\epsilon}_{\mathcal{U}}^{(\varrho)} (C) \triangleq
	\frac{1}{|\mathcal{U}|} \sum_{x\in\mathcal{U}}\bar{\rho}^{(\varrho)}_{C}(x, f_{\mathcal{U}}).
\end{equation}

\noindent
\textbf{Margin Disparity.}
The \textit{margin disparity} \cite{mdd} (denoted as $\Delta$) on a domain $\langle \mathcal{D}, f \rangle$ w.r.t. two classifiers $C, C' \in \mathscr{C}$
is defined as the following:
\begin{equation}
	\Delta_{\mathcal{D}}^{(\varrho)} (C, C') \coloneqq
	\mathbb{E}_{x\sim\mathcal{D}} \big[ \bar{\rho}^{(\varrho)}_{C}(x, C') \big] .
\end{equation}
And its empirical form on a dataset $\mathcal{U}$ is:
\begin{equation}
	\hat{\Delta}_{\mathcal{U}}^{(\varrho)} (C, C') \triangleq
	\frac{1}{|\mathcal{U}|}\sum_{x\in\mathcal{U}}\bar{\rho}_{C}^{(\varrho)}(x, C').
\end{equation}

\noindent
\textbf{Margin Disparity Discrepancy.}
Fixing the threshold $\varrho$,
the \textit{margin disparity discrepancy} \cite{mdd} (denoted as $d^{(\varrho)}$)
for a classifier $C$ in the hypothesis space $ \mathscr{C}$
w.r.t. domain distributions $\mathcal{D}_S$, $\mathcal{D}_T$ is defined as:
\begin{multline}
	d_{C, \mathscr{C}}^{(\varrho)} (\mathcal{D}_S, \mathcal{D}_T) \coloneqq \\
	2 \sup_{C' \in \mathscr{C}} \Big(
	\Delta_{\mathcal{D}_S}^{(\varrho)} (C, C') - \Delta_{\mathcal{D}_T}^{(\varrho)} (C, C')
	\Big).
\end{multline}
The empirical MDD for $C$ w.r.t. source and domain datasets $\mathcal{S}, \mathcal{T}$ is estimated as:
\begin{multline}
	\hat{d}_{C, \mathscr{C}}^{(\varrho)} (\mathcal{S}, \mathcal{T}) \triangleq \\
	2 \max_{C' \in \mathscr{C}} \Big(
	\hat{\Delta}_{\mathcal{S}}^{(\varrho)} (C, C') - \hat{\Delta}_{\mathcal{T}}^{(\varrho)} (C, C')
	\Big).
\end{multline}

\noindent
\textbf{Error Bound w.r.t. MDD.}
With the equations above,
we are able to deliver an \textit{error boundary} of $C$ on $\mathcal{D}_T$, 
with probability at least $1-3\delta$ $(\delta>0)$:
\begin{equation}
	\label{eqn:errbd2}
	\begin{split}
		\epsilon_{\mathcal{D}_T}(C) \le \hat{\epsilon}_{\mathcal{S}}^{(\varrho)}(C)
		&+ \hat{d}_{C, \mathscr{C}}^{(\varrho)} (\mathcal{S}, \mathcal{T})
		+ \lambda_{(\varrho, \mathscr{C}, \mathcal{D}_S, \mathcal{D}_T)} \\
		&+ 2\sqrt{\frac{\log\frac{2}{\delta}}{2|\mathcal{S}|}} + \sqrt{\frac{\log\frac{2}{\delta}}{2|\mathcal{T}|}} \\
		&+ K \Big(|\mathcal{Y}|, \frac{1}{\varrho}, \frac{1}{\sqrt{|\mathcal{S}|}}, \frac{1}{\sqrt{|\mathcal{T}|}} \Big),
	\end{split}
\end{equation}
where $\lambda$ is a constant,
while $K$ is a term positively related to the number of classes $|\mathcal{Y}|$
and negatively related to the margin threshold $\varrho$,
and the sizes of both source and target datasets.

\noindent
\textbf{Hypothesis Neighborhood (MDD).}
As mentioned in Section \ref{sec:hypo_nbhd},
the concept of unchangeable hypothesis class $ \mathscr{C}$ is questionable for the optimization process.
In the scenario of MDD, the optimal hypothesis neighborhood $ \mathscr{C}^*$ is a region
where the local discrepancy ${d}_{C, \mathscr{C}^*}^{(\varrho)} (\mathcal{D}_S, \mathcal{D}_T)$ is the minimum across the universe:
\begin{equation}
	\mathscr{C}^* \coloneqq
	\arg\min_{ \mathscr{C}' \subset \mathscr{C} }
	\big[ \max_{ C \in \mathscr{C}' }
	{d}_{ C, \mathscr{C}' }^{(\varrho)} (\mathcal{D}_S, \mathcal{D}_T)
	\big].
\end{equation}
Similarly, we use $\mathscr{C}$ to denote the hypothesis neighborhood of the current classifier $C$.

\subsection{Cross-Entropy Divergence}
The \textit{cross-entropy loss}, $\mathcal{L}_{\text{CEL}}$, is a pervasive loss function.
For a set of logit vectors $\mathcal{P}$ with ground-truth labels $\mathcal{Q}$:
\begin{equation}
	\mathcal{L}_{\text{CEL}} (\mathcal{P}, \mathcal{Q}) =
	\mathbb{E}_{(X, y)\in(\mathcal{P}, \mathcal{Q})} \big[
	- \log [ \sigma_y (X) ]
	\big],
\end{equation}
where $\sigma_y$ is the softmax possibility for label $y$.
However, the cross-entropy loss cannot measure the divergence between two discrete distribution vectors.

Moreover, although other statistical metrics
(\textit{e.g.} KL-divergence, JS-divergence, etc.) provide decent discrepancy measurements,
their quantitative outputs are usually too insignificant to balance the source supervision losses
calculated via the cross-entropy criterion, which is unfavorable for training CNNs.

To address this disadvantage,
we generalize the cross-entropy loss to composite labels by
defining the \textit{cross-entropy divergence}, $\mathcal{L}_{\text{CED}}$:
\begin{equation}
	\mathcal{L}_{\text{CED}} (\mathcal{P}, \mathcal{Q}) =
	\mathbb{E}_{(X, Y)\in(\mathcal{P}, \mathcal{Q})} \big[
	- Y^{T} \log [ \sigma(X) ]
	\big],
\end{equation}
where $\mathcal{P}$ is the set of predicted logits,
and $\mathcal{Q}$ is the set of composite labels.

\subsection{Algorithm}

The commented SAF-mixup algorithm is displayed in Algorithm \ref{algo:saf_mixup}.

\begin{algorithm}[t]
	\LinesNumbered
	\SetAlgoLined
	\SetAlgoLongEnd
	\DontPrintSemicolon
	\SetKwInput{KwModule}{Module}
	\SetKw{KwBegin}{BEGIN:}
	\SetKw{KwEnd}{END.}
	\SetKwComment{algocmt}{\# }{}
	\newcommand\algocmtfont[1]{\textcolor{blue}{#1}}
	\SetCommentSty{algocmtfont}
	
	\caption{SAF-mixup\label{algo:saf_mixup}}
	
	\KwModule{
		bottleneck $B$, classifier $C$ \\ \hspace{41pt}
		SAF bottlenecks $S_1$, $S_2$ \\ \hspace{41pt}
		SAF weight estimator $S_{\eta}$
	}
	\KwIn{the set of target features $\mathcal{F}_T$}
	\KwOut{augmented features ($\widetilde{\mathcal{F}}_T$, $\widetilde{\mathcal{Y}}_{\mathcal{T}}$)}
	
	\KwBegin \;
	\algocmt*[h]{initialize output sets} \;
	$\widetilde{\mathcal{F}}_T \gets \emptyset$;\: $\widetilde{\mathcal{Y}}_{\mathcal{T}} \gets \emptyset$ \;
	\While(){$\mathcal{F}_T \neq \emptyset$}{
		\algocmt*[h]{draw from $\mathcal{F}_T$ without replacement} \;
		$\phi_1, \phi_2 \gets \operatorname{RandomDrawPair} (\mathcal{F}_T)$ \;
		\algocmt*[h]{feed $\phi_1, \phi_2$ into $S_1, S_2$, respectively;}\;
		\algocmt*[h]{then feed the sum into $S_\eta$ to get weight} \;
		$\eta \gets S_{\eta}\big(S_1(\phi_1) + S_2(\phi_2)\big)$ \;
		\algocmt*[h]{linearly combine $\phi_1, \phi_2$ w.r.t. $\eta$} \;
		$\tilde{\phi} \gets \eta \phi_1 + (1-\eta)\phi_2$;\;
		\algocmt*[h]{get pseudo-labels for the pair} \;
		$\hat{y}_1 \gets C \circ B (\phi_1);\: \hat{y}_2 \gets C \circ B (\phi_2)$ \;
		\algocmt*[h]{similar for $\hat{y}_1, \hat{y}_2$}\;
		$\tilde{y} \gets \eta \hat{y}_1 + (1-\eta)\hat{y}_2$ \;
		\algocmt*[h]{update $\widetilde{\mathcal{F}}_T$ and $\widetilde{\mathcal{Y}}_{\mathcal{T}}$} \;
		$\widetilde{\mathcal{F}}_T \gets \widetilde{\mathcal{F}}_T \cup \{\tilde{\phi}\}$;\:
		$\widetilde{\mathcal{Y}}_{\mathcal{T}} \gets \widetilde{\mathcal{Y}}_{\mathcal{T}} \cup \{\tilde{y}\}$ \;
	}
	\KwEnd
\end{algorithm}

The training scheme for the complete SAF framework is shown in Algorithm \ref{algo:saf}.

\subsection{Implementation Details}

\subsubsection{Network Structures}

The network structure designed for experiments is built upon source codes
of MDD\footnote{\url{https://github.com/thuml/MDD/}} \cite{mdd} and
of ImA\footnote{\url{https://github.com/xiangdal/implicit_alignment/}} \cite{impali}.
The source code can be found in our supplementary materials.

\noindent
\textbf{The feature extractor} $F$ is a ResNet50 \cite{resnet} pre-trained on the ImageNet \cite{imagenet},
following the commonly-used UDA training protocol \cite{dann,cdan}.

\noindent
\textbf{The bottleneck} $B$ uses the following structure:
\begin{itemize}[topsep=2pt, itemsep=2pt, parsep=0pt]
	\item Fully-Connected Layer (2048$\to$1024)
	\item Batch Normalization Layer
	\item ReLU Layer
	\item Dropout Layer (50\%)
\end{itemize}
to filter input features.

\noindent
\textbf{The classifier} $C$ and \textbf{the adversarial module} $D$ shares the same architecture in the MDD structure,
where $D$ is able to regularize $F$ for extraction of transferable as well as discriminative features:
\begin{itemize}[topsep=2pt, itemsep=2pt, parsep=0pt]
	\item Fully-Connected Layer (1024$\to$1024)
	\item ReLU Layer
	\item Dropout Layer (50\%)
	\item Fully-Connected Layer (1024$\to$$|\mathcal{Y}|$)
\end{itemize}
where $|\mathcal{Y}|$ is the number of classes.

\noindent
\textbf{SAF bottlenecks} $S_1, S_2$ use relatively simple structure to process feature representations from $F$:
\begin{itemize}[topsep=2pt, itemsep=2pt, parsep=0pt]
	\item Fully-Connected Layer (2048$\to$384)
	\item ReLU Layer
\end{itemize}
while \textbf{the SAF weight estimator} consists of:
\begin{itemize}[topsep=2pt, itemsep=2pt, parsep=0pt]
	\item Fully-Connected Layer (384$\to$1)
	\item Sigmoid Layer
\end{itemize}
which yields a weight $\eta \in (0, 1)$.

\begin{algorithm}[t]
	\LinesNumbered
	\SetAlgoLined
	\SetAlgoLongEnd
	\DontPrintSemicolon
	\SetKwInput{KwModule}{Module}
	\SetKwInput{KwParam}{Parameter}
	\SetKw{KwBegin}{BEGIN:}
	\SetKw{KwEnd}{END.}
	\SetKwComment{algocmt}{\# }{}
	\newcommand\algocmtfont[1]{\textcolor{blue}{#1}}
	\SetCommentSty{algocmtfont}
	
	\caption{SAF Training Algorithm\label{algo:saf}}
	
	\KwModule{
		feature map $F$, bottleneck $B$ \\ \hspace{41pt}
		classifier $C$, adversarial module $D$ \\ \hspace{41pt}
		SAF mixup module $M$
	}
	\KwParam{
		learning rate $\lambda$
	}
	\KwIn{
		source dataset $(\mathcal{S}, \mathcal{Y}_S)$ \\ \hspace{32pt}
		target dataset $(\mathcal{T}, \emptyset)$
	}
	
	\KwBegin \;
	\algocmt*[h]{source label prediction objective} \;
	$\mathcal{P}_S \gets [C \circ B \circ F](\mathcal{S})$ \;
	$\varepsilon_C \gets L_C(\mathcal{P}_S, \mathcal{Y}_S)$ \;
	\algocmt*[h]{adversarial domain adaptation objective} \;
	$\varepsilon_D \gets L_D(F, B, D, C, \mathcal{S}, \mathcal{T})$ \;
	\algocmt*[h]{SAF-supervision objective} \;
	$ \mathcal{F}_T \gets F(\mathcal{T}) $ \;
	$\widetilde{\mathcal{F}}_T, \widetilde{\mathcal{Y}}_T \gets M(\mathcal{F}_T) $ \;
	$\varepsilon_M \gets L_M (\widetilde{\mathcal{F}}_T, \widetilde{\mathcal{Y}}_T) $ \;
	\algocmt*[h]{backpropagation (denoted as $\gets^*$)} \;
	$F, M, B, C \gets^* -\lambda(\varepsilon_C + \lambda_D\varepsilon_D + \lambda_M\varepsilon_M) $ \;
	$D \gets^* \lambda\varepsilon_D $ \;
	\KwEnd
\end{algorithm}

\subsubsection{Hyperparameters}

The GRL \cite{dann} weight $\lambda_D$ is initially set to $0$ and gradually increases to $0.1$,
with the increasing function:
\begin{equation}
	\lambda_D (t) = 0.1 \tanh \frac{10 t}{\mathbf{T}},
\end{equation}
where $t$ is the current iteration number, and $\mathbf{T} = 10^{5}$ is the total training iteration.

Similarly, the SAF mixup weight $\lambda_M$ increases from $0$ to $0.1$ in a slower pace:
\begin{equation}
	\lambda_M (t) = 0.1 \tanh \frac{5t}{\mathbf{T}}.
\end{equation}


\begin{figure*}[h]
	\centering
	\small
	\begin{subfigure}{0.49\linewidth}
		\centering
		\includegraphics[width=\textwidth]{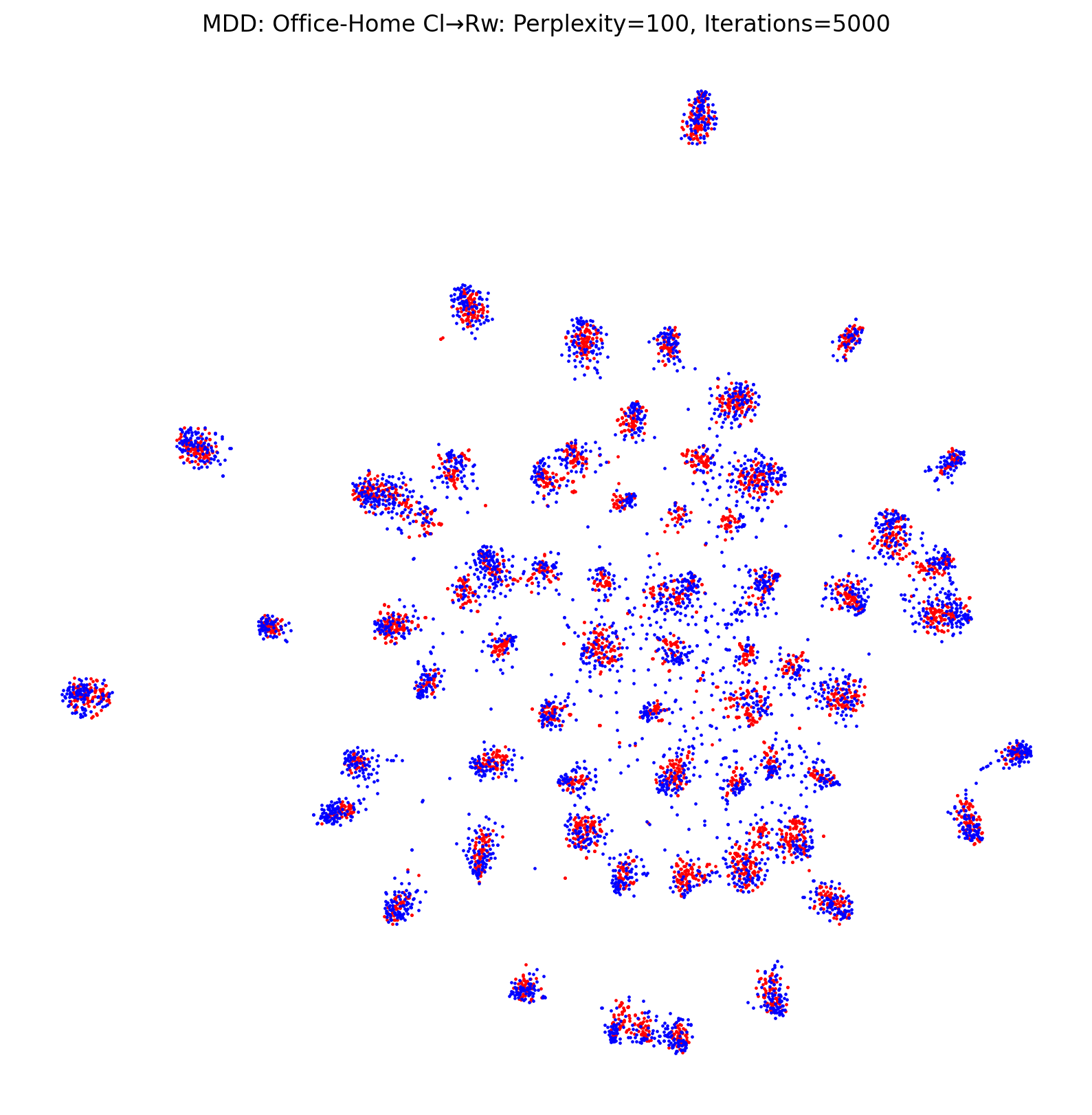}
		\caption{MDD: Cl$\to$Rw}
		\label{fig:tsne_mdd_ohcr_full}
	\end{subfigure}
	\begin{subfigure}{0.49\linewidth}
		\centering
		\includegraphics[width=\textwidth]{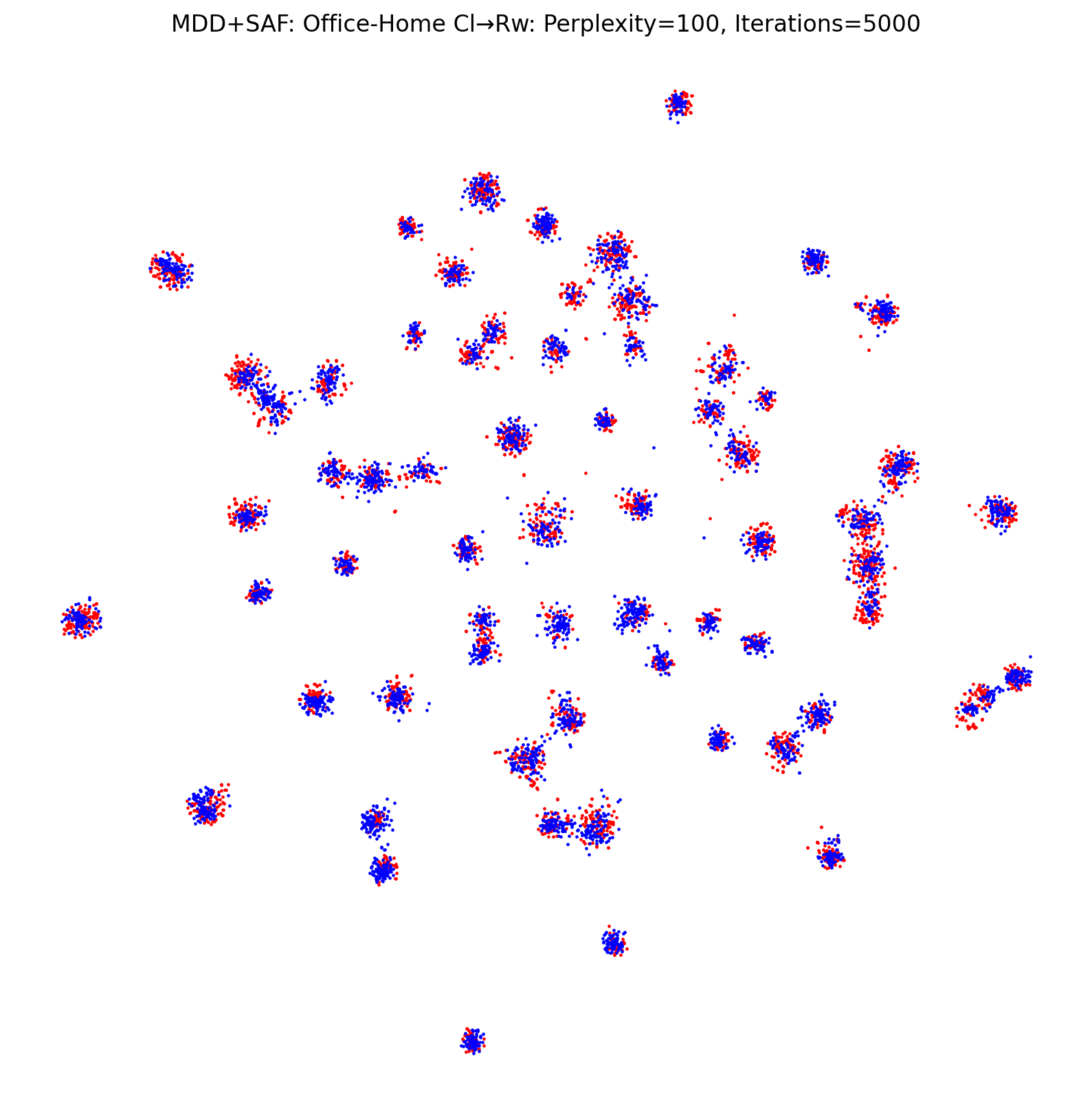}
		\caption{MDD+SAF: Cl$\to$Rw}
		\label{fig:tsne_saf_ohcr_full}
	\end{subfigure}
	
	\begin{subfigure}{0.49\linewidth}
		\centering
		\includegraphics[width=\textwidth]{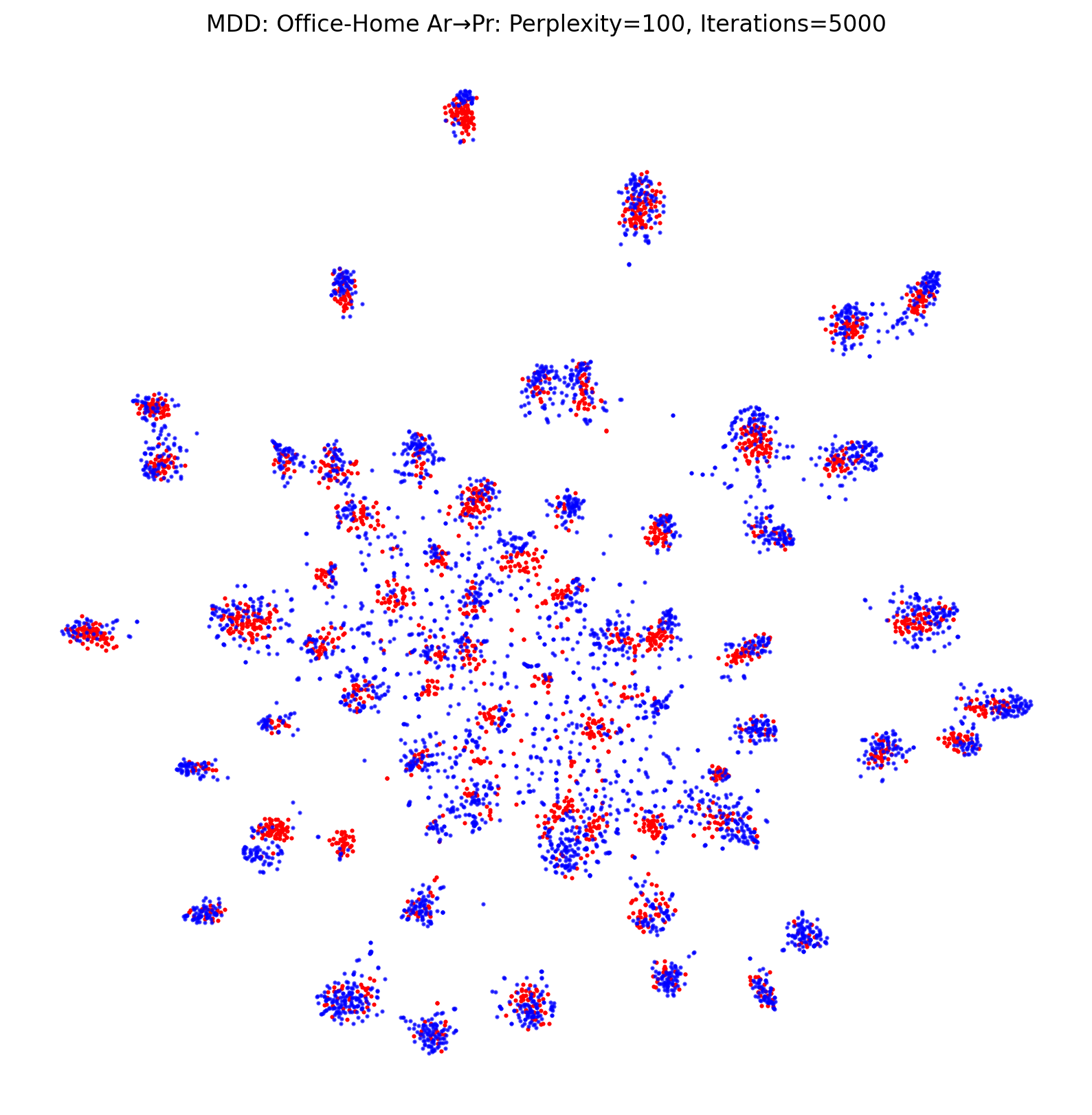}
		\caption{MDD: Ar$\to$Pr}
		\label{fig:tsne_mdd_ohap_full}
	\end{subfigure}
	\begin{subfigure}{0.49\linewidth}
		\centering
		\includegraphics[width=\textwidth]{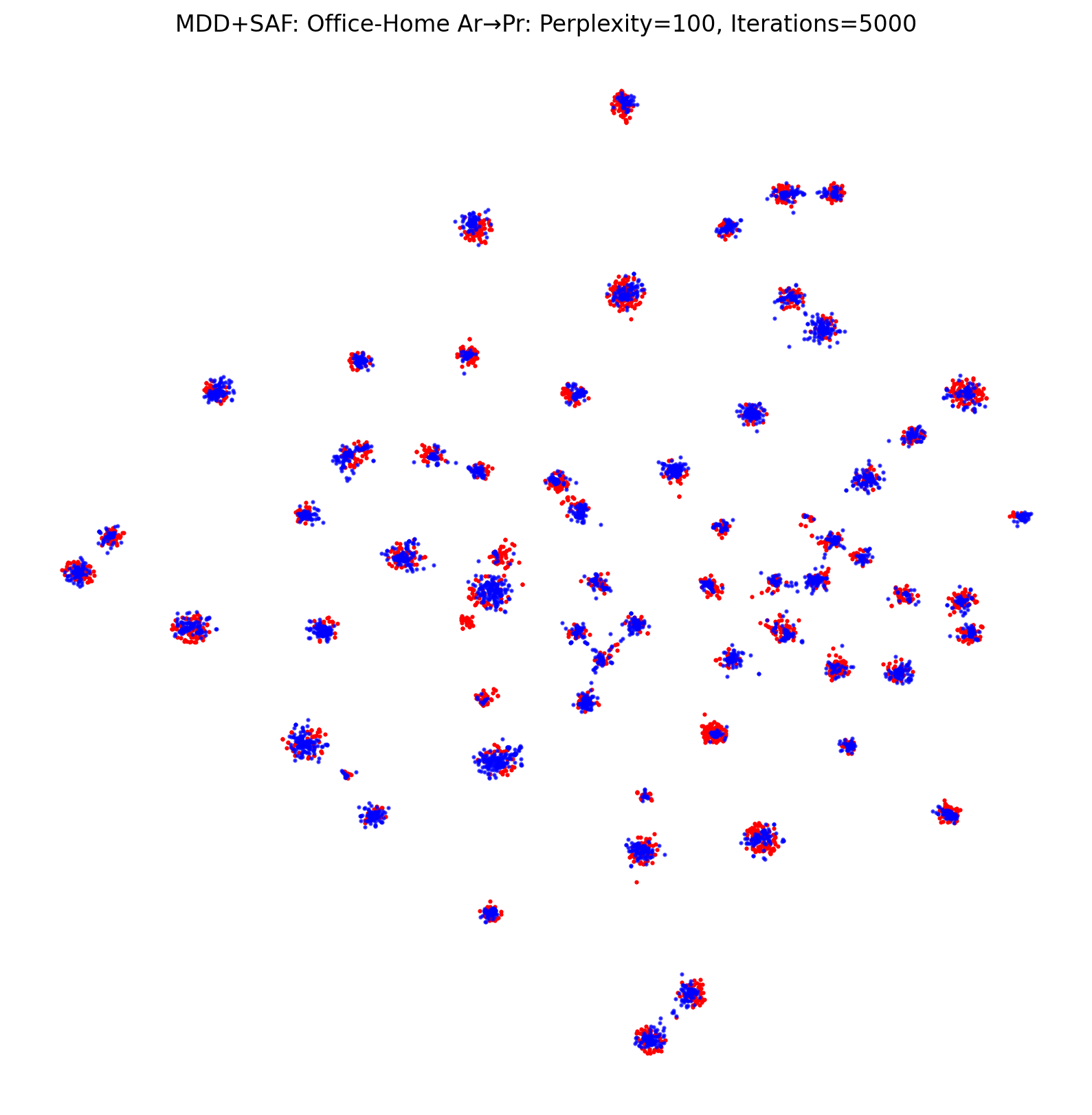}
		\caption{MDD+SAF: Ar$\to$Pr}
		\label{fig:tsne_saf_ohap_full}
	\end{subfigure}
	\caption{
		t-SNE visualizations for features extracted from Office-Home Cl$\to$Rw and Ar$\to$Pr tasks.
		\textbf{Best viewed in color.}
		\textcolor{red}{Red} and \textcolor{blue}{blue} dots represent source and target feature representations, respectively.
	}
	\label{fig:tsne_full}
\end{figure*}

\subsection{Visualizations}

Two sets of t-SNE visualizations \cite{tsne} are shown in Figure \ref{fig:tsne_full}.
Figure \ref{fig:tsne_mdd_ohcr_full} and \ref{fig:tsne_saf_ohcr_full} displays the features extracted from the Office-Home Cl$\to$Rw task.
Figure \ref{fig:tsne_mdd_ohap} and \ref{fig:tsne_saf_ohap_full} displays the features extracted from the Office-Home Ar$\to$Pr task.

The visualization comparison between MDD and SAF shows that the SAF framework
greatly improves feature clustering and semantic transferring between two domains.

\end{document}